\documentclass[10pt,final,twocolumn,journal]{IEEEtran}
\usepackage{fancyhdr}
\usepackage{subfigure}
\usepackage{setspace}
\usepackage{cite}      
\usepackage{graphicx}  
\usepackage{fancyhdr}
\usepackage{amssymb,amsmath}

\begin{document}
\title{People Counting in Crowded and Outdoor Scenes using a Hybrid Multi-Camera Approach}
\author{Fabio Dittrich, 
        Luiz E. S. de Oliveira,
        Alceu S. Britto Jr.
        and~Alessandro~L.~Koerich
\thanks{Fabio Dittrich is with Laboratoire d'Imagerie, Vision et Intelligence Artificielle, Ecole de Technologie Superieure, 1100 Notre-Dame West H3A 1E3, Montreal, QC, Canada. E-mail:fabio.dittrich@livia.etsmtl.ca}
\thanks{Luiz E. S. de Oliveira is with the Federal University of Paran\'{a}, Department of Informatics, Centro Polit\'{e}cnico, CP19011 Curitiba PR 81531-970 Brazil E-mail:lesoliveira@inf.ufpr.br}
\thanks{Alceu S. Britto Jr. is with Pontifical Catholic University of Paran\'{a}, Postgraduate Program in Computer Science, R. Imaculada Concei\c{c}\~{a}o, 1155 Curitiba PR 80215-901 Brazil E-mail:alceu@ppgia.pucpr.br}
\thanks{Alessandro L. Koerich is with \'Ecole de Technologie Sup\'erieure, Department of Software and IT Engineering, 1100 Notre-Dame West H3A 1E3, Montreal, QC, Canada. E-mail:alessandro.koerich@etsmtl.ca}
\thanks{August 2015.}}

\maketitle
\thispagestyle{plain}

\fancypagestyle{plain}{
\fancyhf{}	
\fancyfoot[L]{}
\fancyfoot[C]{}
\fancyfoot[R]{}
\renewcommand{\headrulewidth}{0pt}
\renewcommand{\footrulewidth}{0pt}
}

\pagestyle{fancy}{
\fancyhf{}
\fancyfoot[R]{}}
\renewcommand{\headrulewidth}{0pt}
\renewcommand{\footrulewidth}{0pt}

\begin{abstract}
This paper presents two novel approaches for people counting in crowded and open environments that combine the information gathered by multiple views. Multiple camera are used to expand the field of view as well as to mitigate the problem of occlusion that commonly affects the performance of counting methods using single cameras. The first approach is regarded as a direct approach and it attempts to segment and count each individual in the crowd. For such an aim, two head detectors trained with head images are employed:  one based on support vector machines and another based on Adaboost perceptron. The second approach, regarded as an indirect approach employs learning algorithms and statistical analysis on the whole crowd to achieve counting. For such an aim, corner points are extracted from groups of people in a foreground image and computed by a learning algorithm which estimates the number of people in the scene. Both approaches count the number of people on the scene and not only on a given image or video frame of the scene. The experimental results obtained on the benchmark PETS2009 video dataset show that proposed indirect method surpasses other methods with improvements of up to 46.7\% and provides accurate counting results for the crowded scenes. On the other hand, the direct method shows high error rates due to the fact that the latter has much more complex problems to solve, such as segmentation of heads.
\end{abstract}


%

\section{Introduction}
Visual surveillance is a major research area in computer vision. In video-based surveillance systems, the need to detect events and carry out measurements is recurrent \cite{Hochuli2007,Huang2010,Anwar2012}. The most common information to be detected are people entering or leaving specific areas in the scene, people taking or leaving objects in the scene, crowd formation, fights, and the number of people in the scene \cite{Zulkifley2012,Talu2011,Guo2012,Subburaman2012,Morerio2012}. The issue of accurately estimating the number of people in a scene has several real-life applications such as dynamic control of traffic lights to optimize pedestrians flow based on the analysis of the number of people, controlling the entrance of passengers in a subway station, environmental security, city planning, etc. However, it can be really difficult to estimate the number of people using video when the surveillance scenario is not controlled such is the case of outdoors, environments where the lighting changes constantly and/or occlusions occur frequently \cite{Caballero2011}.

In the last years, several researchers have tackled the problem of people counting. However, most of the works have dealt with people counting in controlled environments where the main constraints are related to the camera position, the number of people moving and the circulation areas \cite{Kettnaker1999,Kim2002,Valle2007}. Counting the number of people has been accomplished using direct and indirect approaches \cite{Albiol2009,Chan2009,Ryan2010,Sidla2006,Sharma2009,Zhao2009}. While the direct approach attempts to segment each individual in scene with classifiers, the indirect one makes use of learning algorithms or statistical analysis of the whole crowd in order to achieve the counting result. For instance, Sharma et al. \cite{Sharma2009} present a direct method which uses an algorithm trained with features called edgelets (arms, shoulders, etc. contours from the silhouette of a person). Each person is segmented, tracked through the frames and counted. Other direct approaches for people counting are: blob counting after background subtraction \cite{Kim2002}, omega shape detection \cite{Sidla2006} (the Greek letter $\Omega$ as the shoulders and head contours) and face detection \cite{Zhao2009,Koerich2010,Oliveira2011,Zavaschi2013}, to name a few. However, as crowd density increases in a scene, it might get impossible to count people individually because most of the features used in detection of a single person fail to give an accurate description of the crowd \cite{Morerio2012}. On the other hand, instead of segmenting each person in the scene, the indirect approach described in \cite{Ryan2010} extracts a set of local features from groups of people in a foreground image. The features are area, perimeter, perimeter-area ratio, edges and edge angle histogram. A density map is computed to weight each pixel, compensating for perspective. Each group has its size estimated using a least-squares linear model which uses the extracted features. Finally, the total count is the sum of all group sizes. Examples of other indirect methods are: training with corner points \cite{Albiol2009} and dynamic texture model \cite{Chan2009}.

One of the most recurrent problems in CCTV systems is the occlusion and lack of visibility in crowded and cluttered scenes. If a person is visually isolated it is much simpler to perform the tasks of detection and tracking \cite{Khan2009}. Increasing the number of people in a scene increases the occlusions. Therefore, occlusion and lack of visibility in dense crowded scenes also makes more difficult to count people correctly and consistently. However, these problems are particularly hard to tackle in single camera systems \cite{Khan2006}. The use of multiple cameras becomes necessary when one wishes to detect and count multiple occluding people in a complex environment. Multiview counting intends to decrease the hidden regions and provide 3D information about the people and the scene by making use of redundant information from different viewpoints \cite{Khan2009}. One alternative is to employ two or more views of the same scene by placing cameras at different positions. In such a way, persons that might be missing in one view may be visible in another view. The same approach can be employed to the problem of people counting. However, there is a lack of methods that employ multiple cameras to tackle the problem of people counting. 


Khan and Shah \cite{Khan2009} have present a multiview approach to solve the problem of tracking individual people correctly in crowded and cluttered scenes. In their approach evidence is gathered from all of the cameras into a synergistic framework and detection and tracking results are propagated back to each view. To such an aim a planar homographic occupancy constraint that fuses foreground likelihood information from multiple views is used to resolve occlusions and localize people on a reference scene plane. The robustness of the approach is demonstrated in challenging multiview crowded scenes.

Wu et al. \cite{Wu2009} have proposed a multi-object multi-camera framework for tracking large numbers of tightly-spaced objects that rapidly move in three dimensions. They have used a greedy randomized adaptive search procedure to find correspondences across multiple views. The occlusions are iteratively solved considering a relaxed assignment problem where one-to-one correspondence is relaxed. .After correspondences are established, object trajectories are estimated by stereoscopic reconstruction using an epipolar-neighborhood search. Snidaro et al. \cite{Snidaro2009} have presented a method for fusing the data of two cameras. The detection of and tracking of objects is carried out via classification by an ensemble of classifiers learned online using color histograms, local binary patterns and Haar-like features. The position of the object on a ground plane map is estimated by fusing likelihood maps which are further approximated by Gaussian function.

Verstockt et al. \cite{Verstockt2009} have proposed a multi-view homography-based approach for object location in compressed video surveillance sequences which is able to accurately localize moving objects on a ground plane using multiple camera data. Object detection in single views is carried out using macro block partition followed by blob merging, convex hull citing and noise removal. The objects found in single views are projected onto a ground plane using homography and object locations are extracted by detecting local maxima on the accumulated ground plane image.
  
Nitta et al. \cite{Nitta2013} have presented a method for counting the number of people traveling over a wide area monitored by spatially disjoint multiple cameras with non-overlapping fields of view. The proposed method counts the number of people traversing across each pair of camera€™ fields of view by estimating the flows between the foreground regions which have disappeared from and appeared in the camera views within a short time interval.
 
Pane et al. \cite{Pane2013} have proposed a novel approach for people counting in stores for business analytics using stereo vision based on background suppression and the clustering on the 3D point cloud by means of mean shift with a cylindrical kernel followed by an adult people classifier which exploits a fitness measure with respect to a cylindrical human body model. Experiments were carried out on two real setups and the results have demonstrated the accuracy of the proposed solution.
 
Li et al. \cite{Li2012} have presented a people counting system which estimates the number of people across multiple cameras with partial overlapping fields of views. They use a multi-object tracking method by means of synthesizing the local-feature-level information into object-level based on an electing and weighting mechanism. The counting results from multiple cameras are integrated using a homography transform and similarity measurement rules. Experiments results demonstrate that their system is effective and accurate for multi-camera people counting.
  
This paper deals specifically with the problem of counting people in a crowded and open environment. We are interested in situations where there are too many people in the scene that partial or total occlusions, both static and dynamic, are common and it cannot be guaranteed that any person will be visually isolated at a given frame. The main challenge in crowd image analysis is to design automatic systems for obtaining information about the movements of, and collective behavior of, individuals and groups within a crowded scene while dealing with unpredictable object movements, people or scene appearance variations, non-rigid parts, intra-occlusions, inter-occlusions and environmental occlusions \cite{Marcenaro2012}. We present a hybrid multi-camera approach that combines the information provided by multiple cameras to deal with many of these issues where both direct and indirect counting strategies are evaluated. The indirect counting strategy is based on the corner point detection method proposed by Albiol et al. \cite{Albiol2009}. The direct count strategy is based on the omega shape detection proposed by Sidla et al. \cite{Sidla2006} using support vector machines and Adaboost perceptron classifiers. The novelty of this work is that both types of techniques are improved with the use of multiple views from the scene to mitigate the problems that may occur due to occlusions where the main challenge is to assure the correspondence between the same persons in two or more camera views. Another improvement is the development of a weighting method based on the homographic  transform which takes into account the position (perspective and distance) of the points of interest in relation to the camera position for the indirect approach since the number of points of interest varies regarding the distance from the camera \cite{Conte2010}.


This paper is organized as follows. Section \ref{sec:met} presents the details of the proposed method. The experimental results are presented in Section \ref{sec:exp}. Finally, the conclusions are stated in the last section.

\section{Proposed Method}
\label{sec:met}
\begin{figure}[t]
    \centering
        \includegraphics[width=90mm]{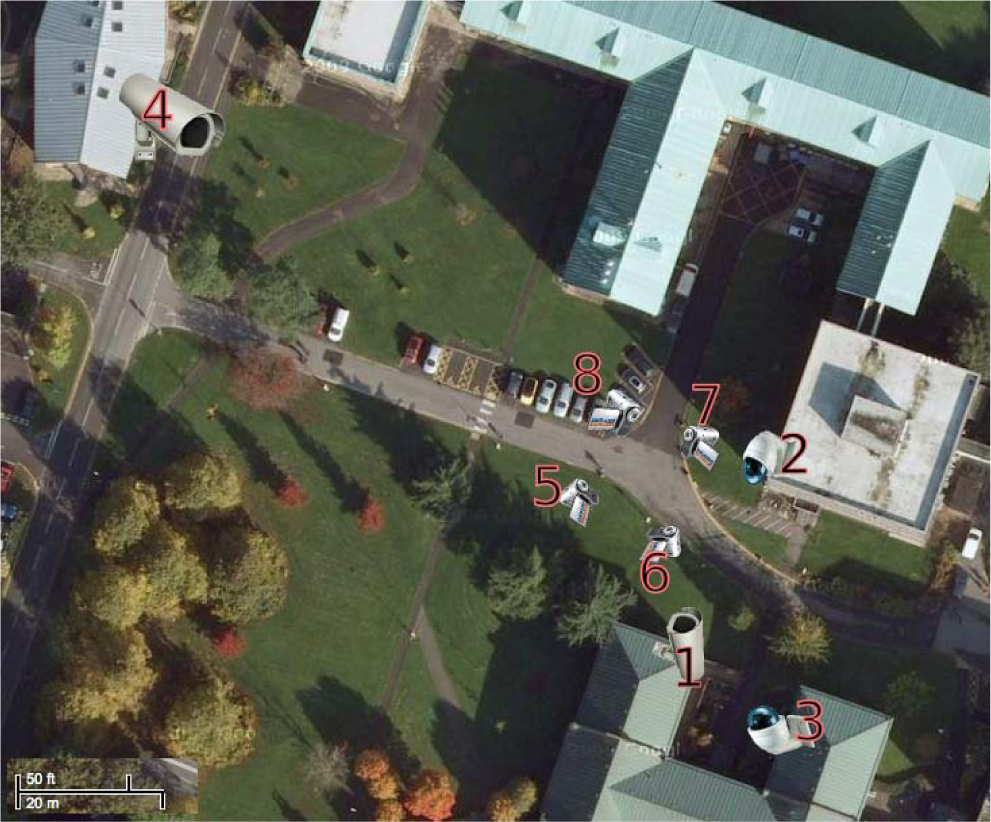}
    \caption{Plan view showing the location and direction of the 8 cameras \cite{Ferryman2009}.}
    \label{fig:environment}
\end{figure}

\begin{figure}[t]
    \centering
        \includegraphics[width=90mm]{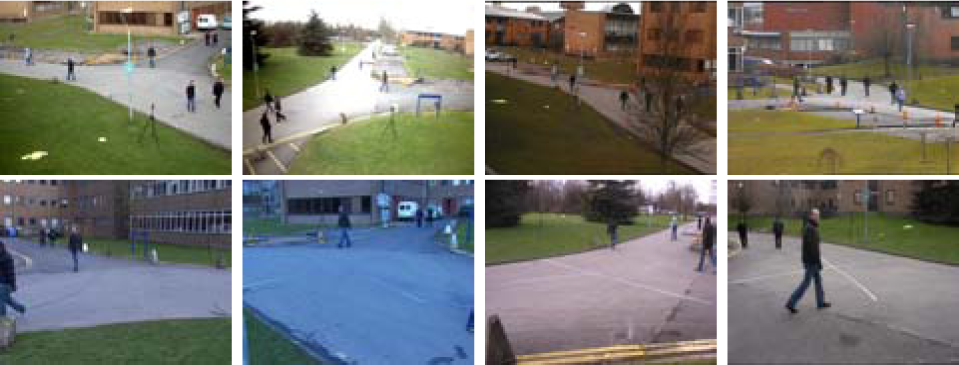}
    \caption{Left-to-right, top-to-bottom: the 8 camera views \cite{Ferryman2009}}
    \label{fig:images}
\end{figure}

\begin{figure}[t]
    \centering
    \subfigure[]{
       \includegraphics[width=20mm]{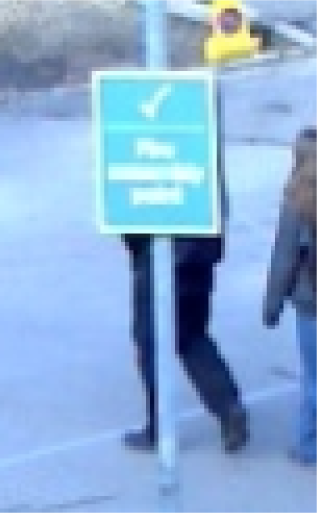}
      \label{fig:statocc}
        \hspace{10mm}
      }
\subfigure[]{
      \includegraphics[width=20mm]{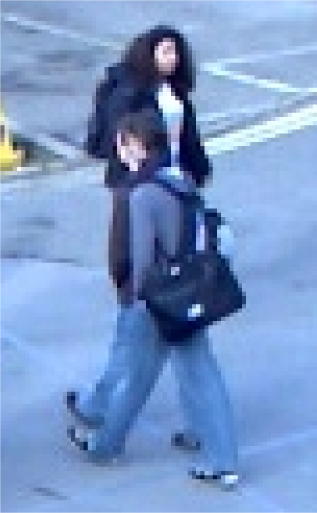}
    \label{fig:dynocc}
   }
   \caption{Example of: (a) static occlusion where almost half of a persons's body is occluded by a sign; (b) dynamic occlusion where two persons are almost totally and partially occluded by other person.}
\end{figure}

In this section we outline our approach to counting people in crowded and outdoor environments. We consider an outdoor environment covered by multiple cameras with partial overlapping of the fields of view as illustrated in Fig. \ref{fig:environment} and samples of the images captured by each camera in Fig. \ref{fig:images}. There are several problems that make counting people in outdoor environments a hard problem: static and dynamic occlusions; different sizes and the perspectives of the people; cameras with different fields of view; changes in the scene illumination. Most of these problems are related to the camera position. Usually, it is very difficult to place a camera straight over the area where people are circulating, which is the best mounting position to avoid occlusions. Furthermore, in open environments, cameras are usually placed to cover a wide area and for this reason they are mounted in oblique positions. The main implication is that the size and the perspective of the people in the scene change considerably according to the distance from the camera as well as static and dynamic occlusions tend to occurs very often. Static occlusion is defined as the obstruction in the view of a person or parts of a person caused by any static object in the scene as shown in Fig. \ref{fig:statocc}. Dynamic occlusion is defined as the obstruction in the view of a person or parts of a person caused by other person or persons as shown in Fig. \ref{fig:dynocc}.

\begin{figure*}[t]
    \centering
        \includegraphics[width=150mm]{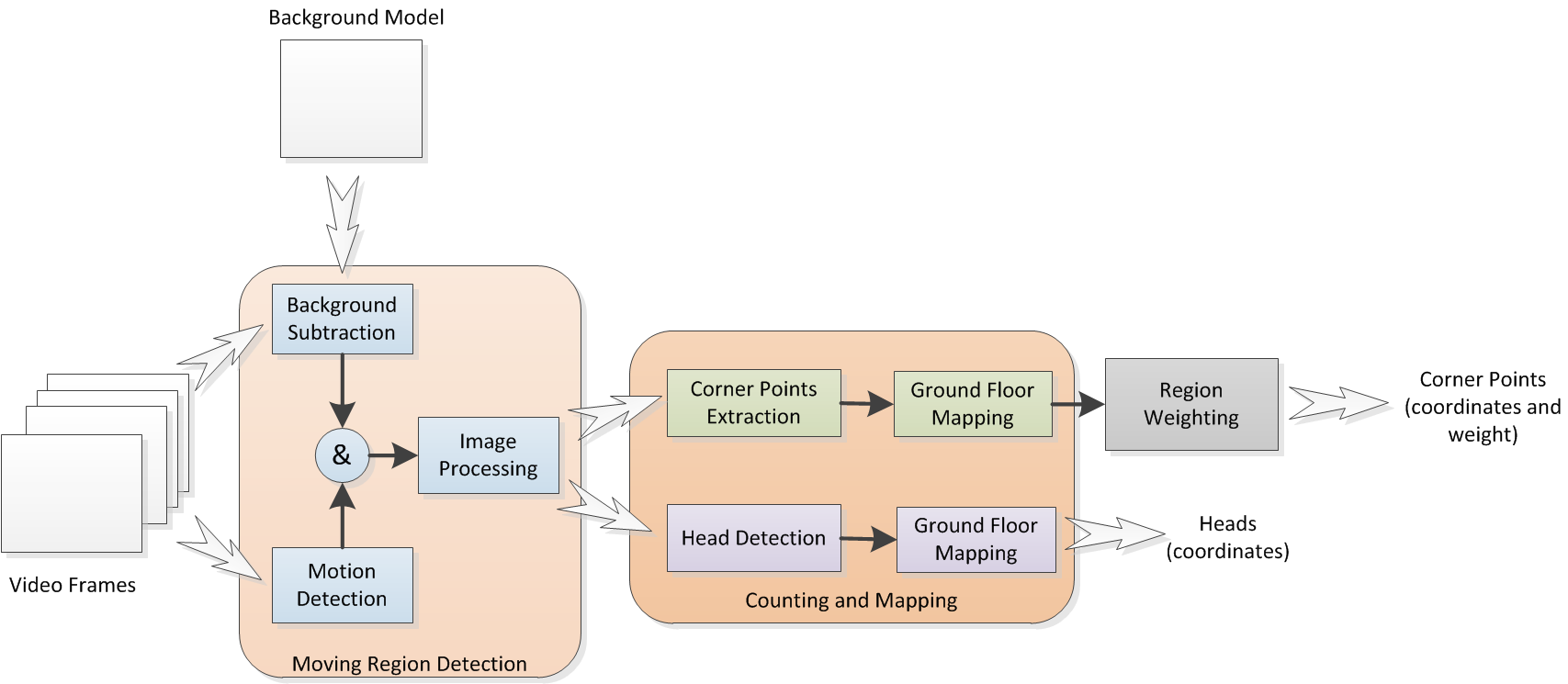}
    \caption{An overview of the proposed method for a single view people counting}
    \label{fig:overview}
\end{figure*}

\begin{figure*}[t]
  \centering
 \subfigure[]{
       \includegraphics[width=90mm]{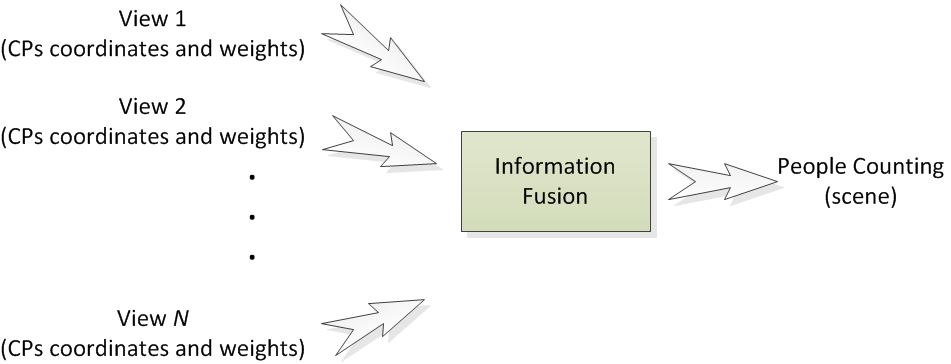}
      \label{fig:cpfusion}
      }
      \hspace{5mm}
\subfigure[]{
      \includegraphics[width=80mm]{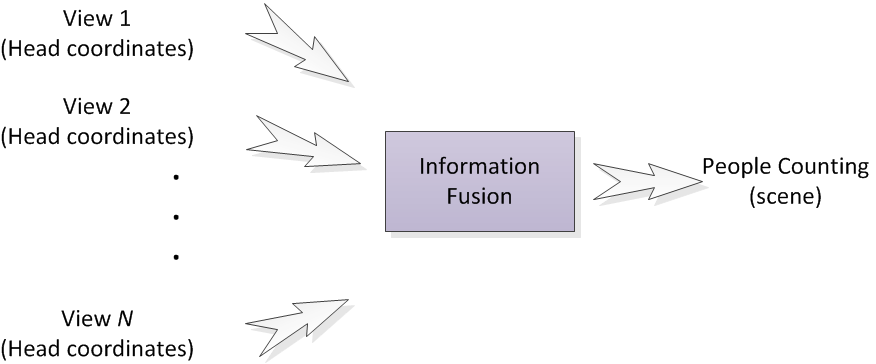}
    \label{fig:headfusion}
   }
   \caption{Fusion of single-view information: (a) coordinates and respective weights of the corner points (CPs) are combined into the ground plane; (b) coordinates of heads are combined into the ground plane.}
\end{figure*}

An overview of the proposed approach is presented in Fig. \ref{fig:overview}. The first step consists in detecting and segmenting the regions of interest, said the regions where people are circulating. We have as input, the video frames as well as a model of the background of the scene. This moving region detection step is made up of background subtraction, motion detection and image processing to enhance the resulting image. The output of this step are the coordinates of blobs that encompass the regions of movement. Next, we have two independent steps which aim is to extract corner points or to detect heads in the moving regions. The output of these two steps are the coordinates of all corner points and heads which are both mapped to the ground floor through a homograph transformation. Some additional steps are used to account for the different distances of the detected corner points from the camera. All these steps are applied to all available views and at the end a rule-based algorithm is used to fuse the information from the different views as shown in Figs. \ref{fig:cpfusion} and \ref{fig:headfusion}. The details of all these steps are presented as follows.

\subsection{Detection of Regions of Movement}
In the image plane we can distinguish between two elements: background and foreground. The former is the visual plane that appears closest to the viewer, while the latter is the plane perceived furthest from the viewer. We can extend this definition and say that the background is made up by the static elements of a scene while the foreground is made up of the dynamic ones. In our case, people in the scene are in the foreground while buildings, trees, street, etc. are in the background.

\begin{figure*}[t]
    \centering
    \subfigure[]{
       \includegraphics[width=55mm]{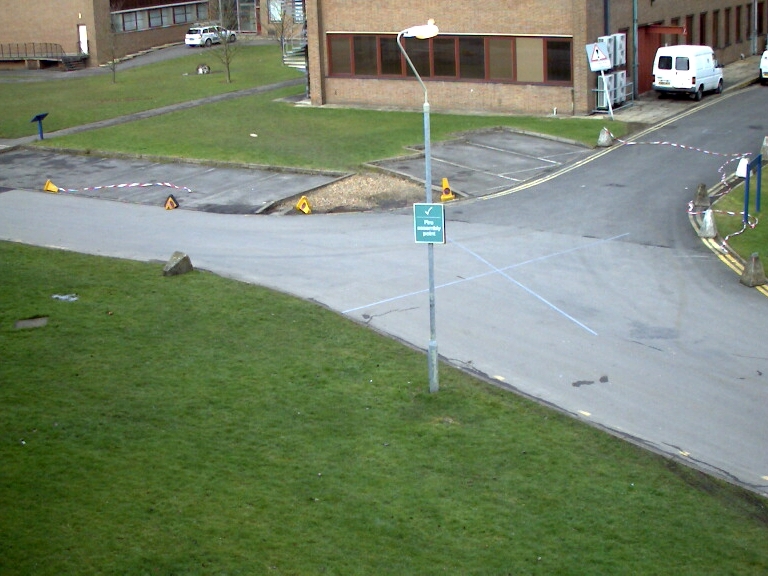}
      \label{fig:back}
      }
\subfigure[]{
      \includegraphics[width=55mm]{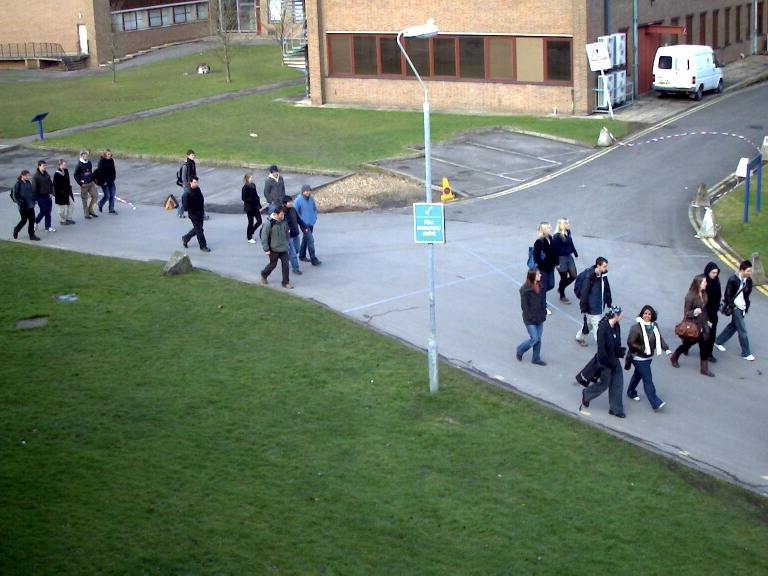}
    \label{fig:frame}
   }
   \subfigure[]{
      \includegraphics[width=55mm]{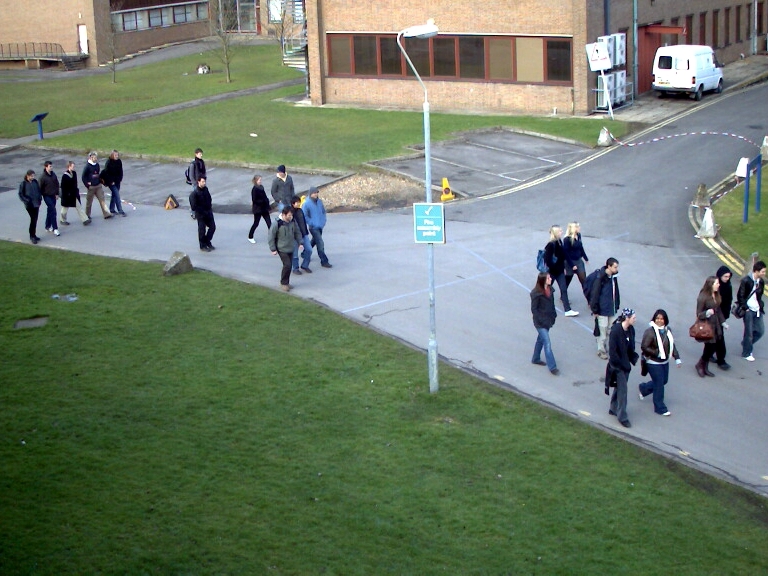}
    \label{fig:frame2}
   }
    \subfigure[]{
      \includegraphics[width=55mm]{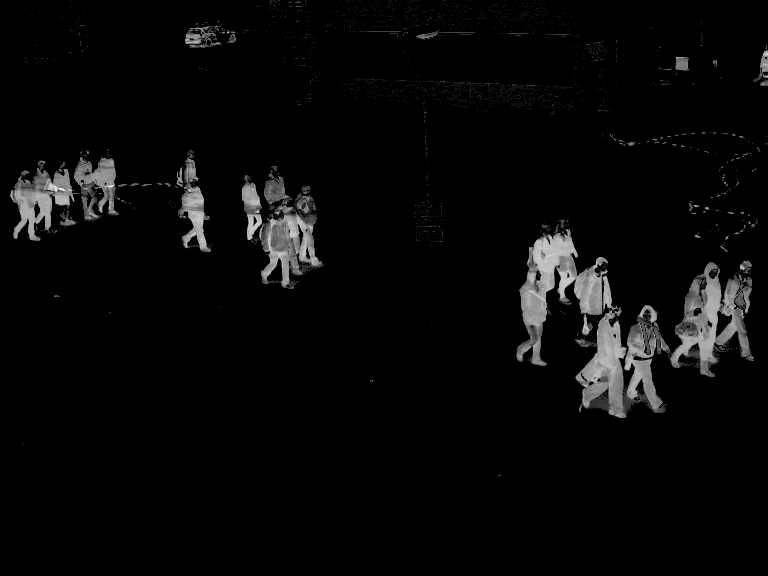}
    \label{fig:backsub}
   }      
    \subfigure[]{
      \includegraphics[width=55mm]{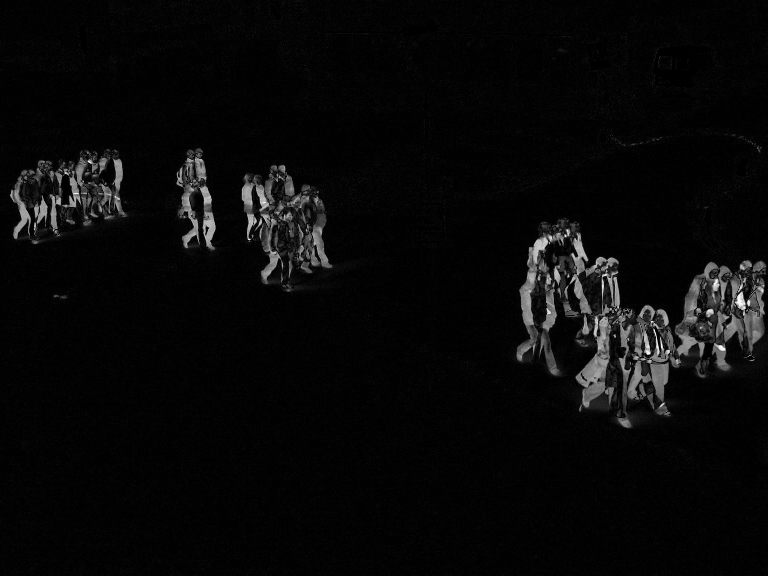}
    \label{fig:subsub}
   }      
    \subfigure[]{
      \includegraphics[width=55mm]{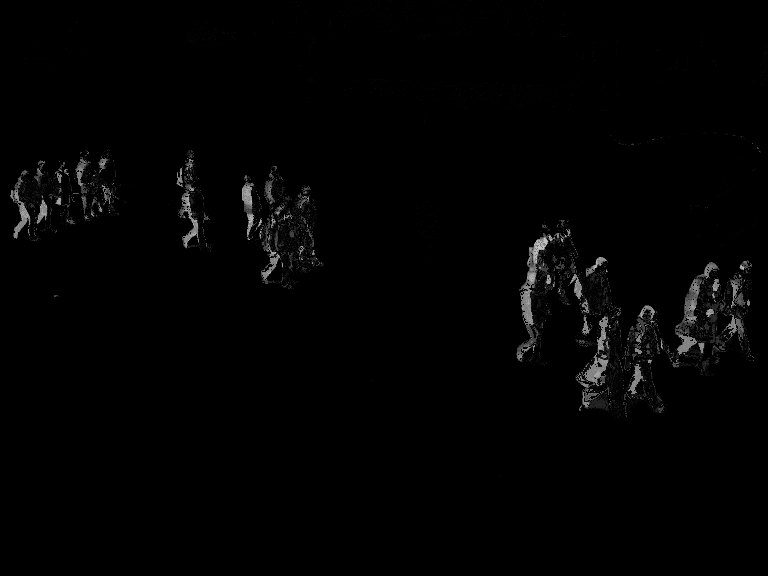}
    \label{fig:And}
   }      
    \subfigure[]{
      \includegraphics[width=55mm]{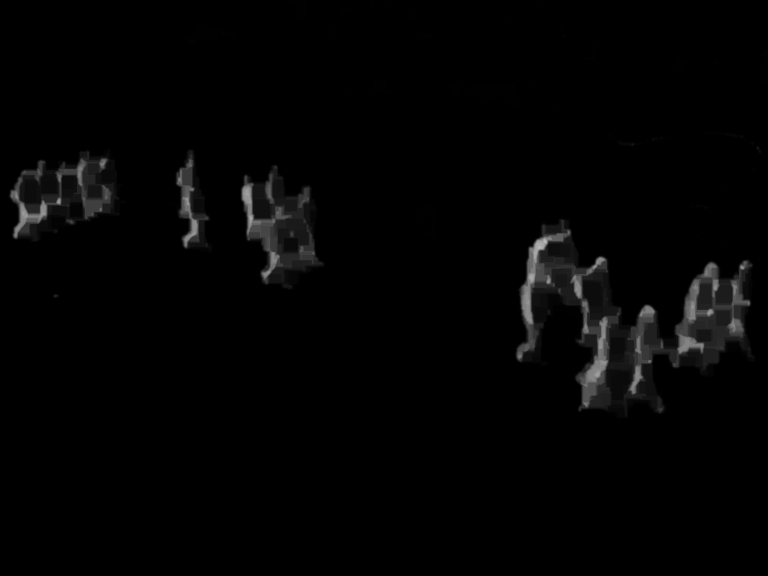}
    \label{fig:closegauss}
   }   
    \subfigure[]{
      \includegraphics[width=55mm]{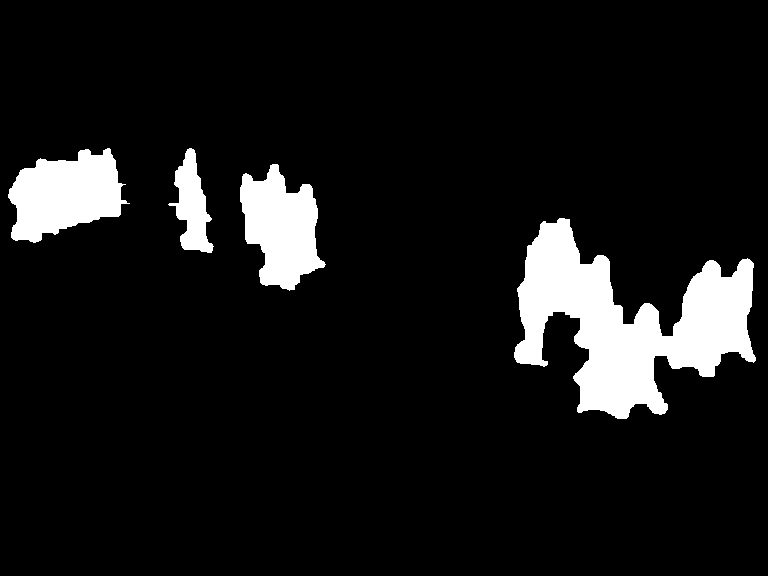}
    \label{fig:hyst}
   }   
    \subfigure[]{
      \includegraphics[width=55mm]{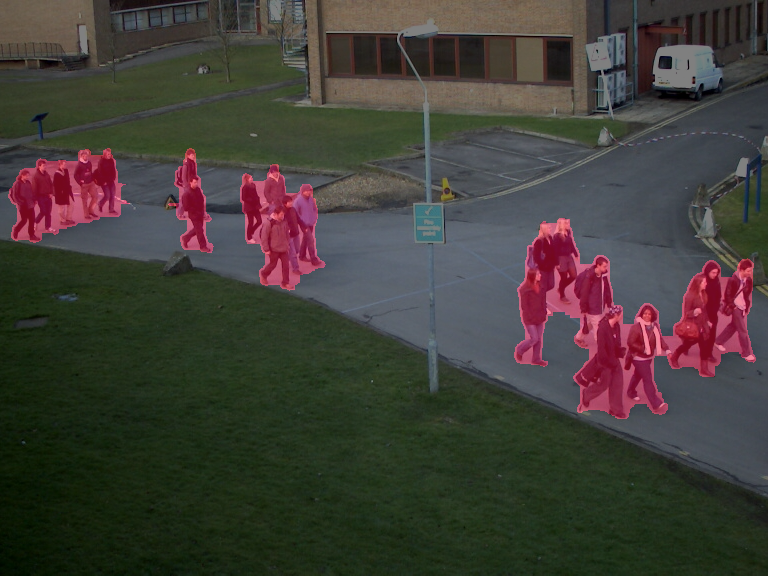}
    \label{fig:blend}
   }
   \caption{(a) An example of a background image; (b) current video frame; (c) subsequent video frame; (d) image resulting from the background subtraction with tolerance; (e) image resulting from the subsequent frame difference; (f) image resulting from the logical AND operation; (g) image after the application of Gaussian blurring and close morphological operator; (h) image resulting of hysteresis thresholding with the regions of movement; (i) blending between the original frame and the regions of movement.}
    \label{fig:preprocess}
\end{figure*}

The first step in determining both the corner points and detecting the heads is to constrain the search space to regions of interest where movements may exist in the scene. For such an aim, a background model is computed by averaging several background images, that is, images that contain only the static elements of the scene (Fig. \ref{fig:back}). All video frames are subtracted, pixel by pixel, from the background model resulting in a difference image that contains only the foreground elements (Fig. \ref{fig:backsub}). However, the resulting image is not perfect since small changes in the scene illumination can produce noises and spurs. Therefore the difference between subsequent video frames is also computed to detect moving regions (Fig. \ref{fig:subsub}). At the end, the results of the background subtraction and motion detection are combined through an logical AND operator (Fig. \ref{fig:And}).

Given a background model denoted as $B(i, j)$ (Fig. \ref{fig:back}) and a current video frame, denoted as $I_t(i, j)$ (Fig. \ref{fig:frame}), the resulting image, denoted as $I_t^\prime(i, j)$ (Fig. \ref{fig:backsub}) is obtained by Eq. \ref{eq:sub}. 

\begin{equation}
I_t^\prime(i, j)=(I_t(i, j)-B(i, j))\wedge (I_t(i, j)-I_{t-1}(i, j))
\label{eq:sub}
\end{equation}

\noindent where $1 \leq i \leq M$ and $1 \leq j \leq N$ are the pixel coordinates, $t$ is the frame index, and $M$ and $N$ are the horizontal (number of rows) and vertical (number of columns) image dimensions respectively.

Next, a Gaussian smoothing operator is used to blur the resulting images to remove detail and noise. For such an aim, an integer-valued kernel (Fig. \ref{fig:closegauss}) that approximates a Gaussian with a standard deviation equal 1.0 is convolved with the image.

\begin{figure}[t]
    \centering
        \includegraphics[width=40mm]{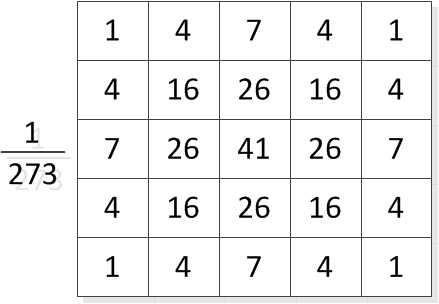}
    \caption{Discrete approximation to Gaussian function with standard deviation equal 1.0}
    \label{fig:gauss}
\end{figure}

The closing morphological operator is used to eliminate any remaining noise and to smooth the contours of the resulting image. The closing operation is the dilation of the imga resulting from the Gaussian blurring, denoted as $I^{\prime\prime} _t(i, j)$, by structuring element $w(k, l)$ followed by erosion by the structuring element, as denoted in Eq. \ref{eq:close}, where $\bullet$ is the closing operator and $I^{\prime\prime\prime}_t(i, j)$ is the image resulting from the closing operation. The closing operation smoothes the contour of the image, fuses narrow breaks, eliminates small holes, and fills gaps in the contour (Fig. \ref{fig:closegauss}).

\begin{equation}
I^{\prime\prime\prime}_t(i, j)=I^{\prime\prime}_t(i, j)\bullet w(k, l)=[g_t(i, j)\oplus w(k, l)]\ominus w(k, l)
\label{eq:close}
\end{equation}

\begin{figure}[t]
    \centering
        \includegraphics[width=20mm]{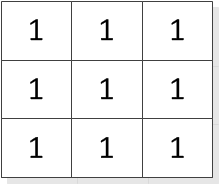}
    \caption{3$\times$3 structural element used in opening and closing morphological operations}
    \label{fig:struct}
\end{figure}

The 3x3 structuring element used in the morphological operation is shown in Fig. \ref{fig:struct}. The structuring element has its origin at the central pixel and it is shifted over the image and at each pixel of the image its elements are compared with the set of the underlying pixels. If the two sets of elements match the condition defined by the set operator, the pixel underneath the origin of the structuring element is set to a pre-defined value.

Finally, the hysteresis thresholding is applied to the resulting image and we regions of movement are summarized with blobs that bound the contours of the moving regions (Fig. \ref{fig:hyst}). Both the corner point and the head detection are only computed within these bounding regions. This reduces the false detection of corner points and heads as well as speeds up the process. Fig. \ref{fig:blend} shows a blending of the detected regions of movement and the current video frame.



\subsection{Corner Point Extraction}
The extraction of the corner points is based on the work of Albiol et al. \cite{Albiol2009} which employs the Harris corner detector \cite{Harris1988}. The Harris corner detector determines the average change of image intensity from a small shifting window.  Let $I_s(i,j)$ be a sub-image of a moving region and $E(u,v)$ be the difference in intensity for a displacement of $(u,v)$ in all directions as computed by Eq. \ref{eq:harris1}.
\begin{equation}
E(u,v) = \sum_{i,j} z(i,j)[I_s(i+u,j+v)-I_s(i,j)]^2
\label{eq:harris1}
\end{equation}
\noindent where $z(i,j)$ is a window function, $I_s(i+u,j+v)$ is the shifted intensity and $I_s(i,j)$ is the intensity of the sub-image. The window function $z(i,j) = exp(-(i^2+j^2)/(2\sigma^2))$ is a Gaussian kernel that gives weights to pixels underneath. Here we use the discrete approximation to Gaussian function with standard deviation equal 1.0 (Fig. \ref{fig:gauss}) which is convolved with the subtraction of the shifted intensity and intensity.

\begin{figure}[t]
    \centering
    \subfigure[]{
       \includegraphics[width=20mm]{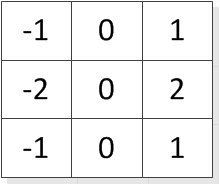}
      \label{fig:Gx}
      }
\subfigure[]{
      \includegraphics[width=20mm]{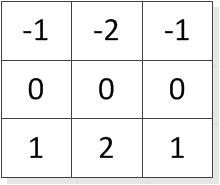}
    \label{fig:Gy}
   }
   \caption{3$\times$3 Sobel kernels: (a) horizontal direction; (b)  vertical direction.}
\end{figure}

We have to maximize the second term of Eq. \ref{eq:harris1} for corner detection. Applying Taylor expansion and some mathematical steps, $E(u,v)$ can be approximated by Eq. \ref{eq:harris2}, where $G_x(i, j)$ and $G_y(i, j)$ are image derivatives in the $i$ and $j$ directions respectively and they are computed using a Sobel operator. The horizontal and vertical derivatives are computed by convolving the sub-image $I_s(i,j)$ with two $3\times 3$ kernels as shown in Fig. \ref{fig:Gx} and  \ref{fig:Gy}.
\begin{equation}
E(u,v) \approx [u,v] M \left[
\begin{array}{l}
u\\
v\\
\end{array}
\right]
\label{eq:harris2}
\end{equation}

\noindent where

\begin{equation}
M = \sum_{i,j}z(i,j) \left[
\begin{array}{ll}
G_x^2(i, j) & G_x(i, j)G_y(i, j)\\
G_x(i, j)G_y(i, j) & G_y^2(i, j)\\
\end{array}
\right]
\label{eq:harris3}
\end{equation}

After that, the eigenvalues of $M$ are computed and used to decided whether a region is corner, edge or flat. Assuming that $\lambda_1, \lambda_2$ are the non-sorted eigenvalues of $M$, we have a corner region when $\lambda_1$ and  $\lambda_2$ are large, a edge when $\lambda_1 \ll \lambda_2$ or $\lambda_2 \ll \lambda_1$, and a flat region when $\lambda_1$ and  $\lambda_2$ are small. Let $\lambda_{min}(i,j)$ and $\lambda_{max}(i,j)$ be the smallest and largest eigenvalue of $M$ respectively, the discriminant function proposed by Albiol et al \cite{Albiol2009} in Eq. \ref{eq:discr} is used to decide whether we have a corner point or not.

\begin{equation}
 D(i, j) = \frac{\lambda_{min}(i, j)}{\lambda_{max}(i, j)}
 \label{eq:discr}
\end{equation}

Furthermore, $D(i,j)$ is retained as a corner point if it is a local maximum and if $D(i,j) > Th_D$ and $G(i,j) > Th_G$ where $Th_D$ and $Th_G$ are two thresholds that are set empirically. Besides these constraints there is an additional constraint to guarantee a minimum distance between corner points. For such an aim squared and circular masks are centered in each corner point and if other corner points falls inside the mask, only the corner point with the highest value of $D(i,j)$ is retained.

Once all corner points have been extracted, a homography transformation is applied to map them to the correspondent points into the ground plane view.

\subsection{Head Detection}
An alternative approach to count people in a crowded scene is through the detection of head regions. Due to the top-oblique position of the cameras, the head is the most visible part of the body in a crowded scene, and the assumption is that the top of the person's body is less susceptible to both static and dynamic occlusions. The proposed method is based on supervised binary classifiers trained on head and non-head image samples. The first binary classifier is based on support vector machines (SVM) with a Gaussian kernel trained on normalized raw images, that is, the intensity values of the raw pixels of images convert to grayscale. The second binary classifier is an Adaboost Perceptron trained on Haar-like features extracted from the grayscale raw images \cite{Belaroussi2011} . 


Once the classifiers are trained, given an input image, they provide at the output a head/no-head label. Such classifiers are fed with sub-images captured with a $n\times n$ sliding window from the regions where movement is detected. In summary, the regions of movement are scanned with a $n\times n$ sliding window and one pixel step at both horizontal and vertical directions. Both classifiers use this approach for detecting heads and only the training and classification parts are different as it is detailed in the following subsections.

\subsubsection{Haar Features based Adaboost Perceptron}
\begin{figure}[t]
    \centering
        \includegraphics[width=60mm]{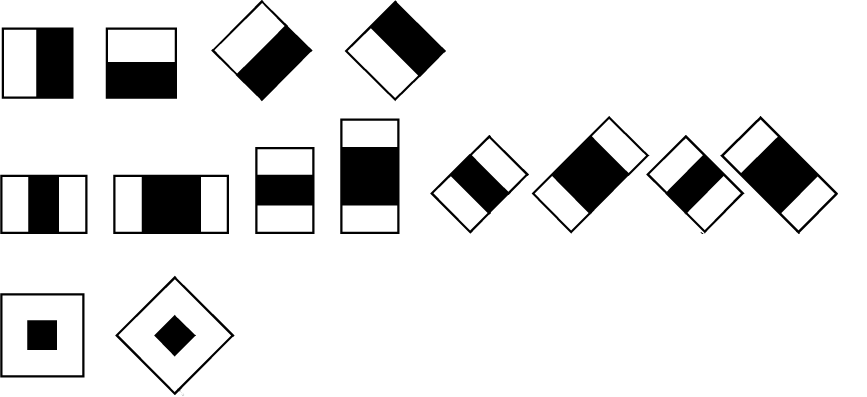}
    \caption{Feature prototypes of edge (first row), line (second row) and center-surround (third row) haar-like features.}
    \label{fig:haar}
\end{figure}

The Haar classifier initially proposed by Papageorgiou et al. \cite{Papageorgiou1998} and improved by Viola and Jones \cite{Viola2001} and Lienhart et al. \cite{Lienhart2002} uses AdaBoost classifier cascades and Haar-like features which accounts for changes in oriented contrast values between regions in the image. There are fourteen feature prototypes as shown in Fig. \ref{fig:haar}, where in the first row we have four edge features, in the second row we have eight line features, and in the last row we have two center-surround features. These prototypes can easily be scaled independently in vertical and horizontal directions to generate a rich, over complete set of features \cite{Lienhart2002}. This allows features to be used to detect heads of various sizes since they may appear close or far from the camera mount position.

We have a specific classifier for each of the multiple views. Therefore, for a given view, a set of positive (heads) and negative (non-head) sample images are used to train a cascade of weak binary classifiers using a variant of AdaBoost algorithm. Such an algorithm allows both the selection of a small number of important features and the training of the cascade of binary classifiers. This is necessary because the fourteen features prototypes can be scaled and this leads to a large number of detected features, for instance, over 180,000 detected features for a $24\times 24$ image \cite{Viola2001}.

Once the cascade of weak classifiers is trained, it receives a video frame with the detected regions of movement and searches for heads of varying sizes by a sliding window that is  moved pixel by pixel over the image at each scale. Starting with the original scale, the features are enlarged until exceeding the size of the image in at least one dimension. Often multiple heads are detect at nearby location and scale at an actual face location. Therefore, multiple nearby detection results are merged \cite{Lienhart2002}.

When a head is detected, the central pixel of the head is projected to the ground plane using the homographic transformation as described in Section \ref{sec:homo}.

\subsubsection{Support Vector Machine Classifier}
\begin{figure*}[t]
    \centering
        \includegraphics[width=150mm]{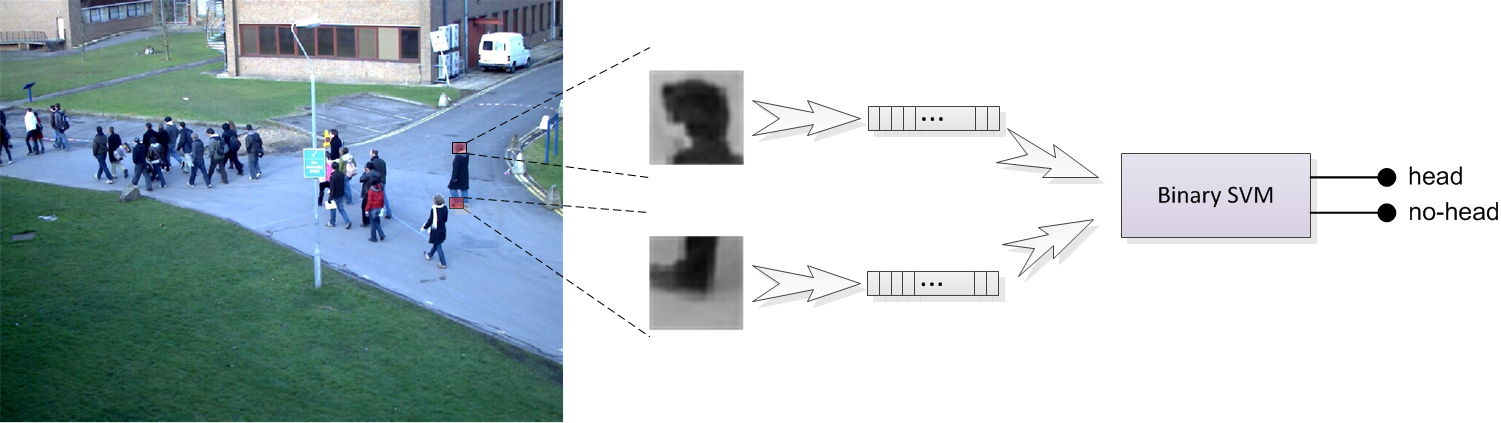}
    \caption{A squared sliding window captures samples of heads and no-heads from the regions of movement which are further normalize and the lines are concatenated to forma feature vector. The features vectors are used to train a binary SVM classifier. }
    \label{fig:svm}
\end{figure*}

The binary SVM classifiers are also trained using a set of positive (heads) and negative (non-head) sample images specific for each view. All images are rescaled to a fixed dimension, said $9\times 9$, and the image rows are concatenated to make up a 81-dimensional feature vector where the features are the raw pixel intensities as show in Fig. \ref{fig:svm}.   


\begin{figure}[t]
    \centering
        \includegraphics[width=60mm]{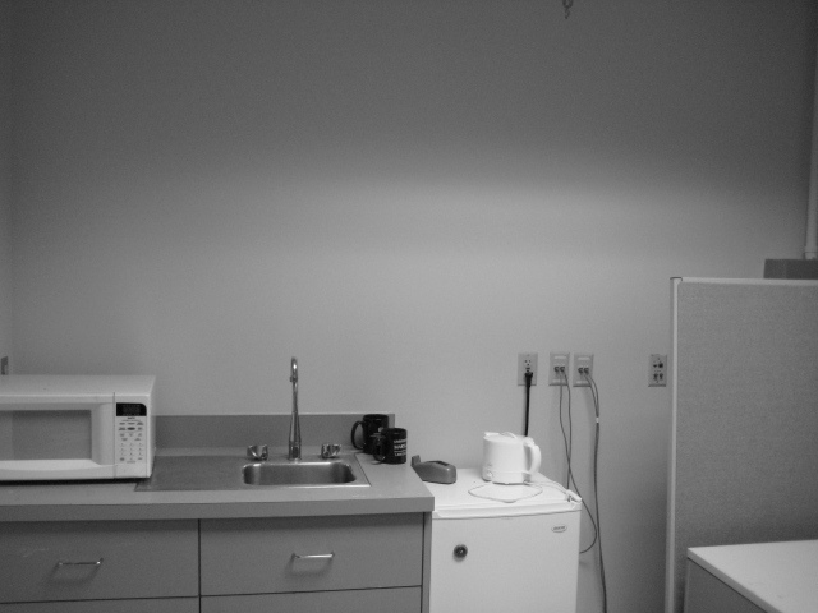}
    \caption{Example of negative image \cite{Seo2012}}
    \label{fig:7}
\end{figure}

During the classification process, for each pixel that belongs to a region of moment, a mask starting with size $9 \times 9$ and scaled up to size $25 \times 25$ centered at this pixel is opened. The raw images are used as features where the gray level intensity of all pixels encompassed by the mask are normalized and summary into a 81-dimensional feature vector. Since this mask slides over all pixels of the regions of movement, multiple heads detected at nearby locations were merged. Finally, the central pixel of the detected heads are projected into the ground plane using the homographic transformation as described in Section \ref{sec:homo}.

\subsection{Homographic Transform}
\label{sec:homo}
The proposed approach for people counting employs multiple partially overlapped views of the scene to count the number of people in the crowd based on the hypothesis that it is possible to reduce the incidence of occlusions because a person which is occluded in one view may not be occluded in other view. Furthermore, with the integration of multiple cameras it is also possible to expand the field of view which is the case of many CCTV surveillance systems. Therefore, it is necessary to make a fusion of all views to put all available information together and to take a decision about the counting. For such an aim the information of the multiple views are projected onto a common base plane using planar homography. The idea is to project all detected corner points and detected heads in the multiple views to a common base plane and use such projected corner points and heads to count the number of persons in the scene. 

Assuming that we have $n$ overlapped views from the same environment which were captured from cameras with different angles and positions as illustrated in Fig. \ref{fig:images}. Let $p_1=(x_1,y_1,1), p_2=(x_2,y_2,1),\dots p_n=(x_n,y_n,1)$ denote the image location of a 3D scene point in views $1, 2,\dots, n$. It is easy to show that the mapping between any two image planes is a homography, independent of the depth of the scene, which is described by a matrix $H$. Therefore, we can look for a set of points in one view and find the corresponding points in other view \cite{Khan2006}.

The homography matrix $H$ is a $3\times 3$ matrix which transforms one plane into another. Given a set of $k$ points of view 1 denoted as ${p_{11}, p_{12},\dots, p_{1k}}$ and the corresponding set of points of view 2 denoted as ${p_{21}, p_{22},\dots, p_{2k}}$, there is a relation between them and the homography matrix $H$, which is given as \cite{Hartley2004}:
  
\begin{equation}
p_{2i}=Hp_{1i}
\end{equation}

Considering the plane points as homogeneous coordinates, that is $p_{1i}=[x_{p_{1i}}, y_{p_{1i}}, z_{p_{1i}}]$ and $p_{2i}=[x_{p_{2i}}, y_{p_{2i}}, z_{p_{2i}}]$, this relation can also be written as:
  
\begin{equation}
\left[ \begin{array}{c} x_{p_{2i}} \\ y_{p_{2i}} \\ 1 \end{array} \right] = \begin{bmatrix} h_{11} & h_{12} & h_{13} \\ h_{21} & h_{22} & h_{23} \\ h_{31} & h_{32} & h_{33}  \end{bmatrix} \times \left[ \begin{array}{c} x_{p_{1i}} \\ y_{p_{1i}} \\ 1 \end{array} \right]
\label{eq:homo}
\end{equation}

\noindent Here we can consider the $z$ coordinates of the plane points as zero since the destination plan is the ground floor and in fact we need to consider only the $x$ and $y$ coordinates of the points in Eq. \ref{eq:homo}. 

\begin {figure} [htbp]
    \centering
        \subfigure[] {\includegraphics[width=43.5mm]{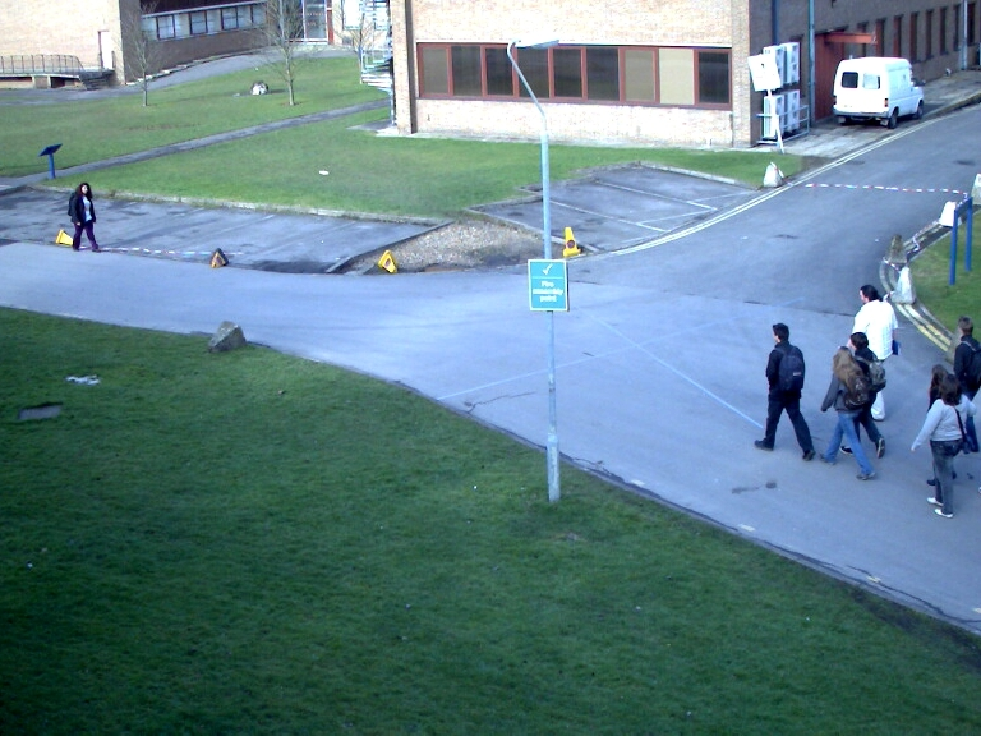}}
        \subfigure[] {\includegraphics[width=43.5mm]{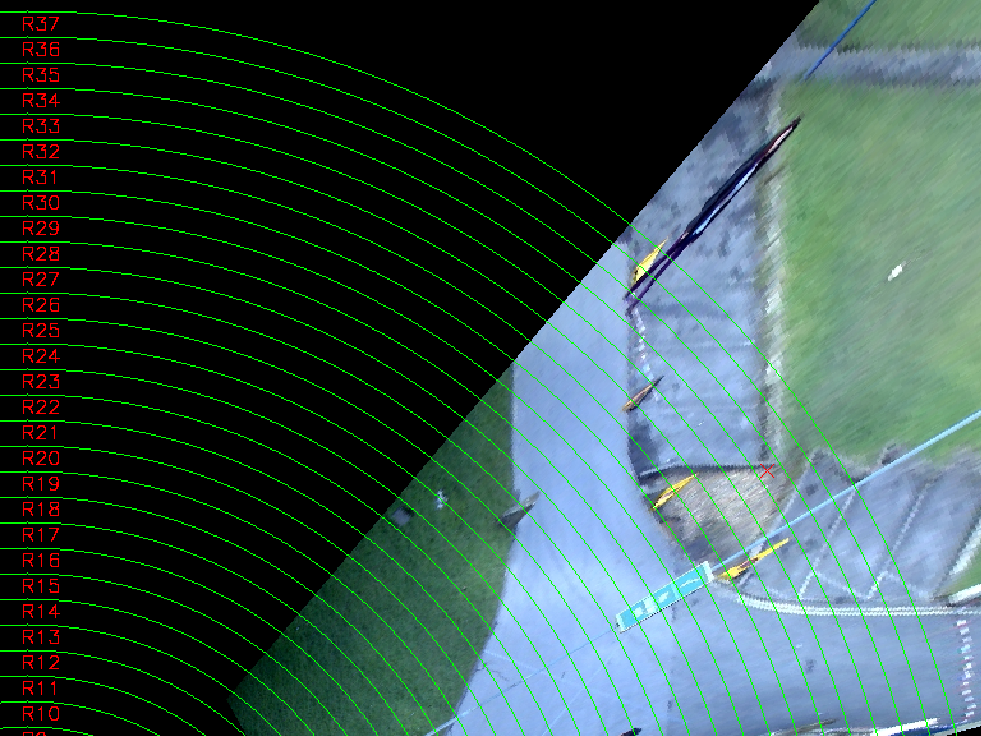}}
    \caption{(a) The original image; (b) the ground plane image with 37 regions created with the homography.}
    \label{fig:3}
\end{figure}

The homographic transform requires the computation of a homographic matrix $H$. This is done using the Tsai camera model \cite{Tsai1986} and the calibration data that is available for each camera. Camera calibration consists in the estimation of a model for an un-calibrated camera. The objective is to find the external parameters (position and orientation relatively to a world co-ordinate system), and the internal parameters of the camera (principal point or image centre, focal length and distortion coefficients). Thus, for each view, several sets of corresponding points are available. The point coordinates that are in the view are in pixels and the point coordinates in the ground floor are in millimeters. To solve this problem, a scale is defined in which one pixel corresponds to a certain amount of millimeters. Now, it is possible to calculate the homography matrix $H$ between the image plane and the ground plane. An illustration of the result of this transformation is shown in Fig. \ref{fig:3}.


The ground plane image has a dimension of $600\times 600$ pixels of height and width. Also, the origin of the scene is centralized on the ground plane image. These procedures were followed to ensure that the camera coordinates fall inside the image.

We end up with an homography matrix for each view relative to the ground plane, said, $H_{v_1\_gp}, H_{v_2\_gp},\dots H_{v_n\_gp}$. The same homography matrices are used project into the ground plane both the corner points and the detected heads. However, when it is assumed that every pixel on the image has its $z$ coordinate equals to zero, projections of all objects onto the ground plane are created. Since not all corner points are on the ground plane, their positions are wrongly mapped when the homography is applied, therefore it is necessary to correct them. 

\subsection{Region Weighting}
\begin {figure} [htbp]
    \centering
        \subfigure[] {\includegraphics[width=43.5mm]{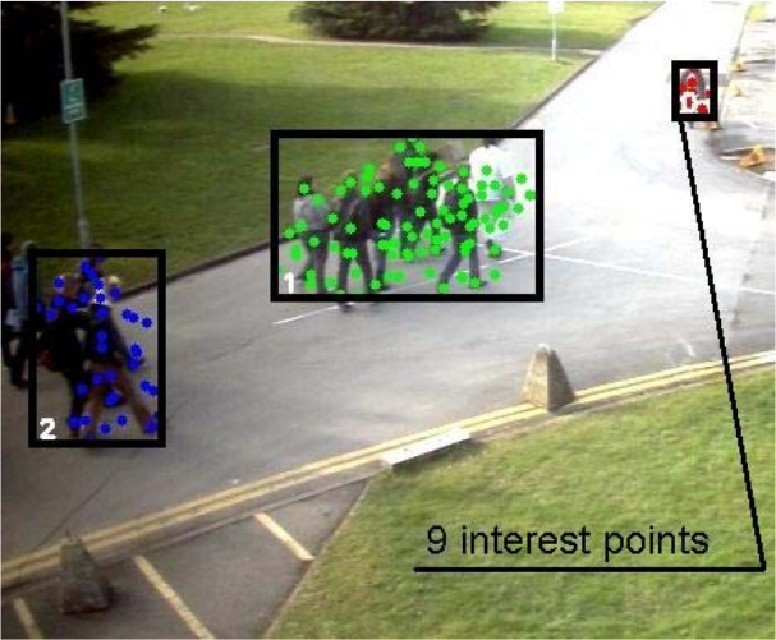}}
        \subfigure[] {\includegraphics[width=43.5mm]{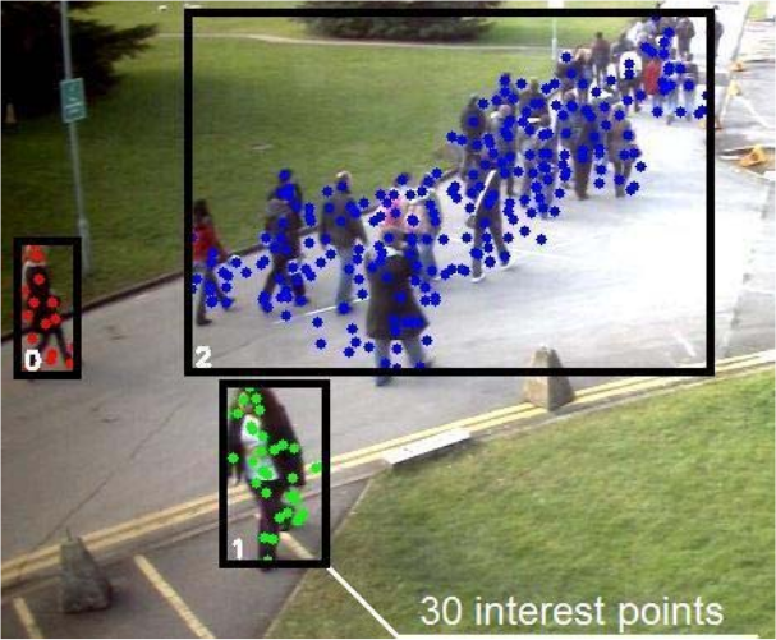}}
    \caption{(a) A person far from the camera has only 9 detected corner points while (b) a person close to the camera has 30 detected corner points.}
    \label{fig:perspec}
\end{figure}

\begin {figure} [htbp]
    \centering
        \subfigure[] {\includegraphics[width=43.5mm]{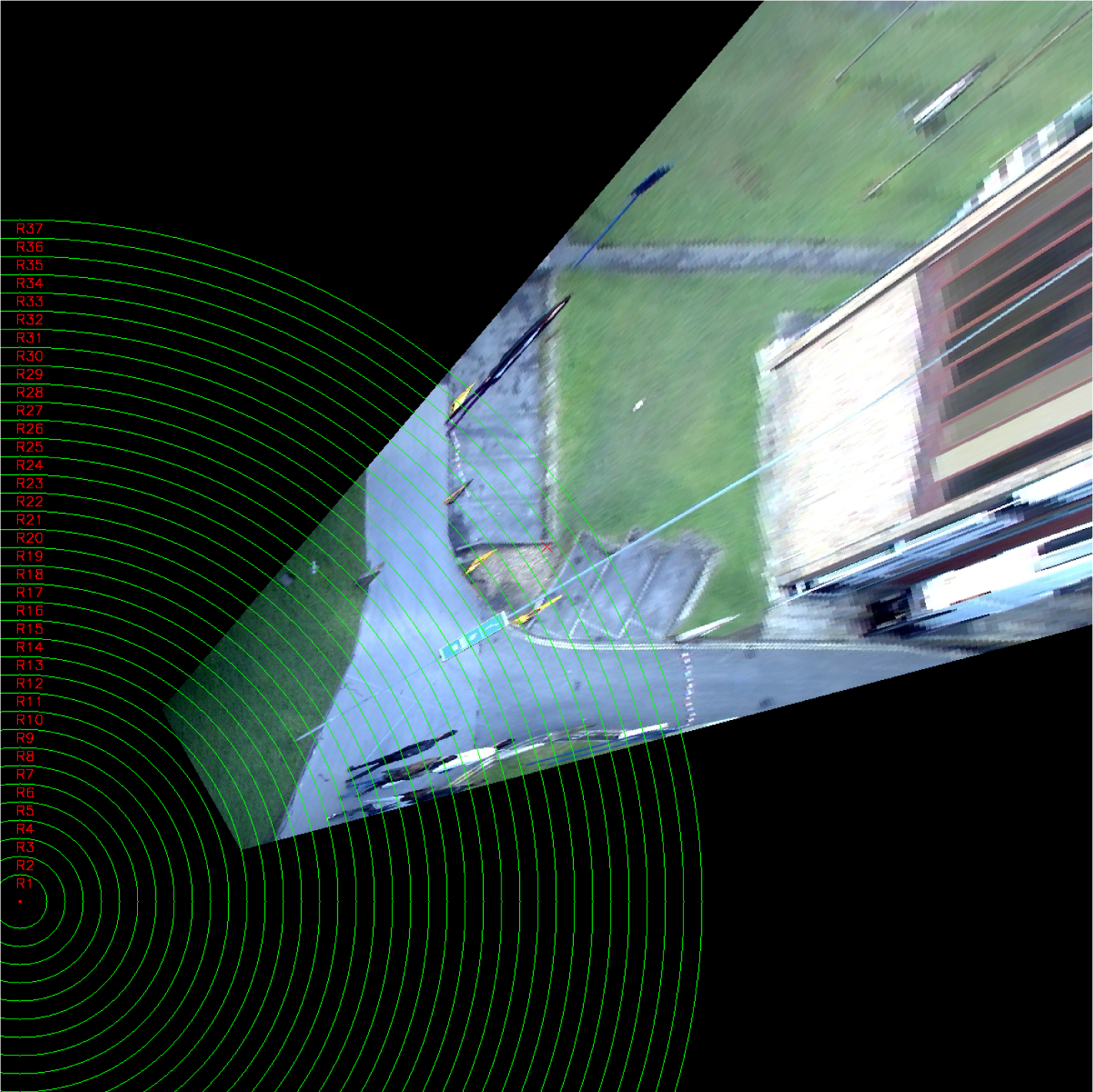}}
        \subfigure[] {\includegraphics[width=43.5mm]{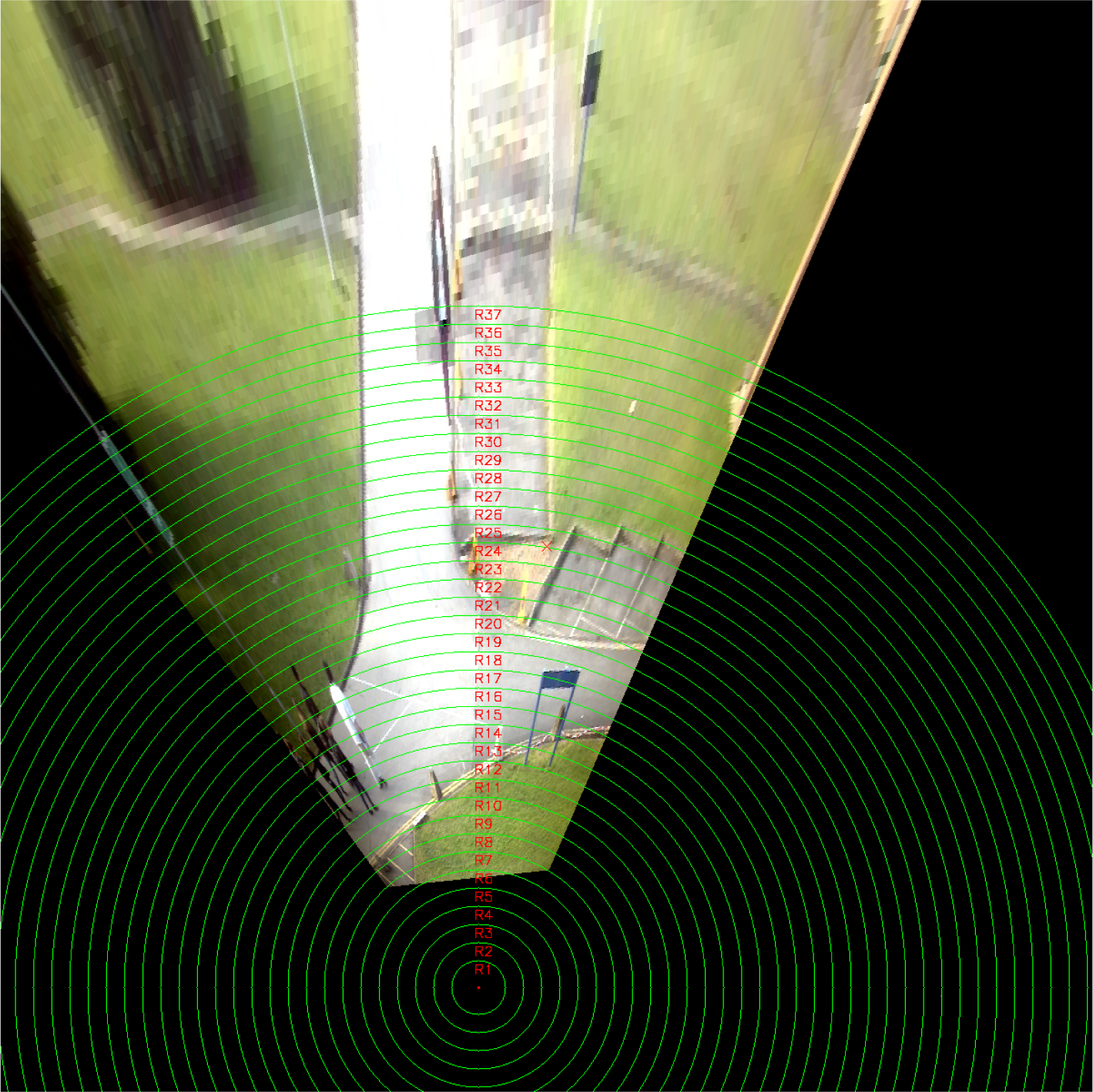}}
    \caption{The 37 circular regions on the ground plane for (a) view 1 and (b) view 2.}
    \label{fig:CircularRegion}
\end{figure}

A major drawback of the Albiol et al. \cite{Albiol2009} method is that it does not consider the perspective and the distance between persons whose corner points are detected and the cameras. This may produce counting errors since a person far from the camera produces less corner points than when the person is closer to the camera, as mentioned by \cite{Conte2010}. Fig. \ref{fig:perspec} clearly shows that the difference in the number of corner points detected is greater than 300\%. For this reason, based on the method proposed by Chan et al. \cite{Chan2009}, we assign weights to each detected corner point according to its distance from the camera. To such an aim, circular regions centered at the camera mounting point have been defined. The size of such regions was determined based on a rule of thumb that a person occupies one meter. With that size, 37 regions are needed to cover the whole path through which the people walk in the scene. Fig. \ref{fig:CircularRegion} shows the 37 regions on the ground plane for two views.

The weights assigned to each of the 37 circular regions we have selected one segment of a video in which a person can be observed without any occlusion since he entered the scene until he left. We have registered, at each frame, his height in pixels and feet position. The feet positions were mapped to the ground plane using the corresponding homographic matrix. Then the average height of the person on each region is computed. The region which contains the system origin ($x=0, y=0, z=0$) received weight $1$ while the others received inversely proportional weights by comparing the average height of each one of them with the region that contains the system origin. For example, if a person at the origin region has an average height of 75 pixels, a closer region where the person average height is 100 pixels will receive a weight equal to 0.75 and so on. Finally, the weights are applied to the detected corner points according to the circular region at the ground plane where they fall after applying the homographic transformation.

\subsection{Projection Correction}
\begin{figure}[t]
    \centering
        \includegraphics[width=39mm]{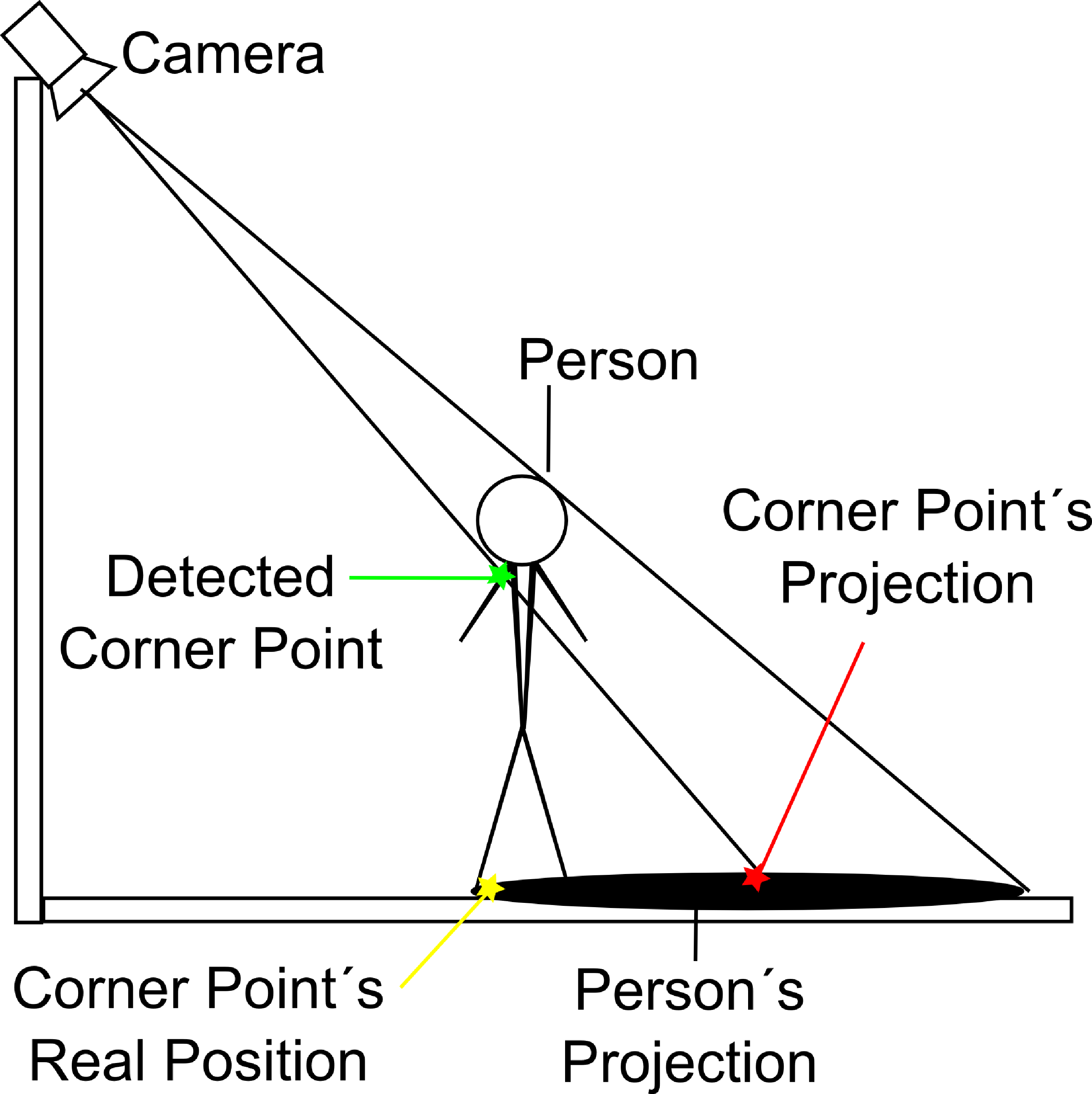}
    \caption{An example of a misprojected detected corner point and its correction by the proposed algorithm.}
    \label{fig:cpprojection}
\end{figure}

The homography is a planar transform and the detected corner points are usually spread through the person's body (see Fig.~\ref{fig:perspec}). When the homographic transform is applied to the corner points, those which are above the ground floor are projected at the wrong places at the ground plane as illustrated in Fig.~\ref{fig:cpprojection}. Therefore, it is necessary to correct the corner's position. To such an aim, the average height of the person in each region that was computed at the region weighting is used. We assume that all persons are represented by an average height and that such an average height correspond to a certain number of pixels, then the following algorithm is used:

\begin{description}
	\item[Step1:  ] For each corner point projected to the ground plane, a straight line is drawn between the projected corner point and the corresponding camera. This is achieved using the Tsai's camera model \cite{Tsai1986};
	\item[Step2:  ] The position of the corner point will be modified by sliding through this line towards the camera as long as:
	\begin{description}
		\item [(a):  ] it is not farther than half of the average height of a person for this region;
		\item [(b):  ] and the corner point position is still inside the circular region.
	\end{description}
\end{description}

\begin{figure}[t]
 \centering
  \includegraphics[width=43.5mm]{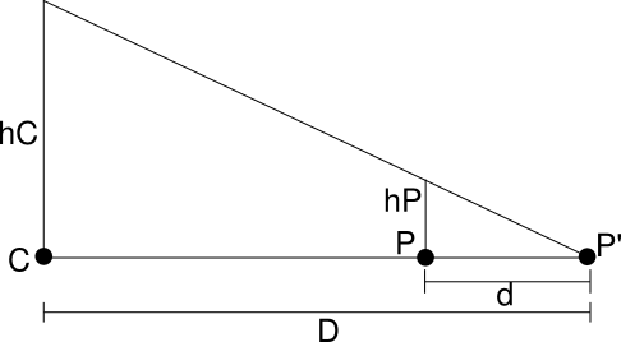}
\caption{The displacement of the head projection at the ground plane ($d$) is calculated using the camera's height ($h_C$), the average person height ($h_P$) and the distance between the point and the camera ($D$).}
\label{fig:5}
\end{figure}

Concerning the detected heads, as they are also above the ground floor, they are projected at the wrong places as well. The procedure to correct the position of the heads projected to the ground plane is similar, but in Step2(a) the head position is modified by displacing it through the straight line is drawn between the projected corner point and the corresponding camera, towards the camera by the full average height of a person for the region. The protection is computed 
using triangle proportion as illustrated in Fig. \ref{fig:5}. Therefore, the displacement of the position of the head into the ground plane, denoted as $d$ is calculated by Eq. \ref{eq:3}. 

\begin{equation}
d = D\frac{h_C}{h_P}
\label{eq:3}
\end{equation}

\noindent where $h_C$ denotes the camera height, $h_P$ denotes de empiric person height, $D$ denotes the Euclidean distance between the corner point and the camera and $d$ the estimated height projection of the corner point.

\subsection{People Counting}
The last step is to establish a relationship between the number of detected corner points and the number of persons in the view. Given the projected corner points and their weights we sum up the number of weighted corner points. We need now to establish a relation between the number of the corner points and the number of persons at a training video frame, for which the ground truth is available. In such a way we obtain the average number of corner points per person at this video frame. Summing up the average number of corner points per person for all frames of the training video and dividing by the number of frames we end up with the average number of corner points per person for a view as given in Eq. \ref{eq:pc}. The same procedure is employed for all available views for which we have training videos with ground truth.

\begin{equation}
\label{eq:pc}
ACPP = \frac{1}{N_f} \sum_{i=1}^{N_f} \frac{ \sum_{j=1}^{N_R} {w_R}_{ij} {N_{CP}}_{ij}}{{N_{GT}}_i}
\end{equation}

\noindent where $ACPP$ denotes the average number of corner points per person for a view, $N_f$ denotes the number of video frames, $N_R$ denotes the number of regions, $w_R$ is the weight for region $j$, $N_{CP}$ is the number of corner points detected at region $j$ at video frame $i$ and ${N_{GT}}_i$ is the truth number of person at the video frame $i$. 



\subsection{Fusion of Information of Multiples Views}
The use of multiple cameras is based on the hypothesis that it is possible to mitigate the problems related to the inter- and intra-occlusions to achieve a more robust estimation of people counting in a scene. Furthermore, the use of multiple cameras also extend the field of view and reduces the problem of environmental occlusions. On the other hand, we have to deal with the data fusion and association problems to make the multiple cameras cooperate for overall scene understanding. The first step towards the fusion of information was the use of the homographic transform to project into a common ground plane the points related to the persons detected in each view \cite{Verstockt2009,Snidaro2009}.

First, let's assume that for each region $r$ of the ground plane, for each video frame $i$ and for each view $v$ we have the number of corner points denoted as ${N_{CP}}_{riv}$, we can combine the ${N_{CP}}_{riv}$'s using simple rules such as maximum (Eq. \ref{eq:max}), minimum (Eq. \ref{eq:min}) or average (Eq. \ref{eq:avg}) to compute the average number of persons in the scene at each frame, denoted as $ANP_i$.

\begin{equation}
\label{eq:max}
ANP_i =\frac{ \sum_{R} \max_{N_v}\{ {N_{CP}}_{ir1}; \dots {N_{CP}}_{irn} \} } {ACPP}
\end{equation}
\begin{equation}
\label{eq:min}
ANP_i =\frac{ \sum_{R} \min_{N_v}\{ {N_{CP}}_{ir1}; \dots {N_{CP}}_{irn} \} } {ACPP}
\end{equation}
\begin{equation}
\label{eq:avg}
ANP_i =\frac{  \frac{1}{N_v}\sum_{R} {N_{CP}}_{ir1} + \dots + {N_{CP}}_{irn}  } {ACPP}
\end{equation}

\noindent where $N_v$ denotes the number of views.

Regarding the detected heads which are represented by the coordinates of the central pixel of the $3\times 3$ detection mask that is overlapped to the head, such a central point is also projected from the views to the ground plane. Once all head points from all views are projected to the ground plane, we need to create a correspondence of points from all views and end up in counting the number of persons in scene. Let's assume that a head point projected to ground plane is denoted as ${ph}_{ivj}$, where $i$ is the video frame index, $v$ is the view, and $j$ is the head point index, for each given point in one view, a $3\times 3$ mask is centered in such a head point and it is scaled up to $n\times n$. If a head point of another view is encompassed by this mask, their mean point is considered the final position of this head point. Assuming that the head detector finds 100\% of the heads in the image, there are three situations in that one head point does not have correspondence to a head point from any other view:

\begin{itemize}
\item the correspondent head was occludes in all other views;
\item this head point is a false detection;
\item this point is in a region of the scene that single view can reach.
\end{itemize}

The last case can be solved by verifying if the point is in one of these regions. If so, it can be counted even without having a correspondence in the other view. However, there is the possibility that a false positive was detected there, which would lead to a false count of a person. This would depend on the false detecting rate of the system. Defining if a false detection or occlusion happened is even more difficult and there is no straightforward answer to this question. Therefore, this approach has many more complex problems to be solved than the corner point detection method. In this work, no solution to this problem was implemented and points without correspondence are simply considered to be false detections, which leads to a high error count.

Finally, once all head points are projected into the ground plane and the correspondences are done, the remaining number of head points lead to the number of persons in the scene at a given video frame.

\section{Experimental Results}
\label{sec:exp}
In this paper we present the experimental results and evaluation for crowd image analysis based on corner point and head counting using the PETS 2009 benchmark dataset. The results achieved by both counting methods proposed in this work were compared to a manually generated ground truth. The ground truth consists in the counting of the number of persons in the scene for each video frame. The number of people in the scene varies between 6 and 42. Since there are some parameters to set such as the average people height, mask/radius size, the ratio number of corner points by the number of people in the view, the combination rules for the corner point method, an other parameters, we have split our dataset into training and testing. Therefore, all parameters adjusted in the training video segments were further evaluated on different video segments. The two proposed methods will be evaluated on this dataset in order to obtain a fair comparison and determine which one of them achieves the best performance in this kind of scenario. The metric used to measure the performance is the average error per frame, denoted as $AepF$ and defined in Eq. \ref{eq:perfmea}

\begin{equation}
Aepf = \frac{1}{nf} \sum_{i=1}^{nf}{n_c}_i - {n_{GT}}_i
\label{eq:perfmea}
\end{equation}

\noindent where $nf$ is the total number of frames in a video, ${n_c}_i$ is the counting provided for frame $i$ by the proposed method and ${n_{GT}}_i$ is the ground truth for frame $i$. The $AepF$ stands for Average Error per Frame, i.e. how many people are miscounted per frame.

The next subsections present the details of the dataset, experiments and results for each counting method.


\subsection{PETS2009 Dataset}
The PETS2009 dataset is a publicly available multi-camera benchmark dataset containing different crowd activities within a real-world environment. The scenarios are filmed from multiple cameras and involve up to approximately forty actors. More specifically, the challenge includes estimation of crowd person count and density, tracking of individual(s) within a crowd, and detection of flow and crowd events. The videos provided on this dataset were captured with cameras located outdoors and far from the walkway through which the crowds pass by. The resolution of the cameras is 768x576 pixels and their average frame rate is 7 frames per second. The scenario is a medium density, regular walking pace crowd flow and the weather is overcast. The definition of crowd is based on a maximum occupancy (100\%) of 40 people in 10 square meters on the ground. One person is assumed to occupy 0.25 square meters on the ground. The unconstrained environment generates a lot of problems such as lighting variation, background movement due to the wind and low resolution.

In particular two video sequences from two views were used the evaluate the proposed method. We have employed the View-001 and View-002 from the video sequences are labelled as Time13-57 and Time13-59 from the dataset L1, source S1, thereinafter S1\_L1\_Time13-57 and S1\_L1\_Time13-59. A manual synchronization was needed because the views were not properly synchronized, which is a requirement of the proposed method. We ended up with 129 and 154 video frames from S1\_L1\_Time13-57 and S1\_L1\_Time13-59 respectively.

The experimental protocol uses a video sequence for training and a different video sequence for testing. The ground truth for the video sequences was built manually by counting both the number of persons in each view for each video frame and the number of persons in the scene, that is, the number of persons that are present in the area covered by View-001 and View-002.




\begin{table*}
 \renewcommand{\arraystretch}{1.3}
 \footnotesize
 \caption{Camera calibration data for View-001 and View-002 respectively}
 \label{tab:calib}
 \begin{center}
 \begin{tabular}{|c|c|c|c|c|c|c|c|}
 \hline
\multicolumn{8}{|c|}{Geometry} \\
\hline
width & height & ncx & nfx & dx & dy & dpx & dpy \\
\hline
768 & 576 & 795 & 752 & 4.850e-03 & 4.650e-03 & 5.1273271277e-03 & 4.650e-03 \\
768 & 576 & 795 & 752 & 4.850e-03 & 4.650e-03 & 5.1273271277e-03 & 4.650e-03\\
\hline
\multicolumn{8}{|c|}{Intrinsic parameters} \\
\hline
f & k & \multicolumn{2}{|c|}{$c_x$} & \multicolumn{2}{|c|}{$c_y$} & \multicolumn{2}{|c|}{$s_x$} \\
\hline
5.5549183034 & 5.1113043639e-03 & \multicolumn{2}{|c|}{3.2422149053e+02} & \multicolumn{2}{|c|}{2.8256650051e+02} & \multicolumn{2}{|c|}{1.0937855397} \\
3.1316979686e & 3.7880262468e-02 & \multicolumn{2}{|c|}{3.6105480377e+02} & \multicolumn{2}{|c|}{3.0732879992e+02} & \multicolumn{2}{|c|}{1.0894268482}\\
\hline
\multicolumn{8}{|c|}{Extrinsic parameters} \\
\hline
tx & ty & \multicolumn{2}{|c|}{tz} & rz & ry & \multicolumn{2}{|c|}{rz} \\
\hline
8.2873214225e+02 & -3.1754796051e+03 & \multicolumn{2}{|c|}{3.5469298547e+04} & 2.0405458695 & -8.9337703748e-01 & \multicolumn{2}{|c|}{-4.3056124791e-01} \\
7.4969800372e+03 & -1.9402122072e+03 & \multicolumn{2}{|c|}{2.3838027680e+04} & 1.8665618542 & 1.5219705811e-01 & \multicolumn{2}{|c|}{4.5968889283e-02} \\
\hline
 \end{tabular}
 \end{center}
 \end{table*}

Tab. \ref{tab:calib} shows the parameters of the Tsai camera model for the cameras used to capture the videos of View-001 and View-002 where $f$ is the focal length of camera, $k$ is the radial lens distortion coefficient, $c_x$ and $c_y$ are the coordinates of centre of radial lens distortion, $s_x$ is a scale factor to account for any uncertainty due to imperfections in hardware timing for scanning and digitization, $r_x$, $r_y$ and $r_z$ are the rotation angles for the transformation between the world and camera coordinates, and $t_x$, $t_y$ and $t_z$ are the translation components for the transformation between the world and camera coordinates \cite{Tsai1986}.

\subsection{Corner Point Results}
\begin{figure}[ht]
\centering
  \includegraphics[width=90mm]{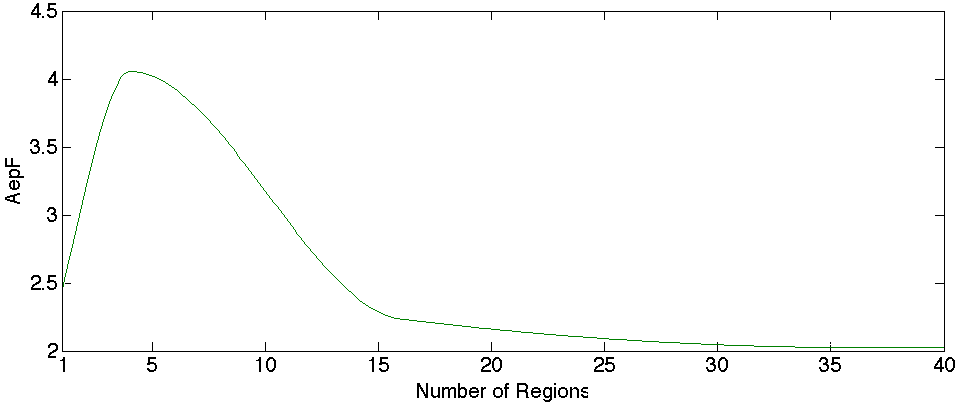}
\caption{Variation of the $AepF$ according to the number of regions for View-002 (S1\_L1\_Time13-57). }
\label{fig:RXAepF}
\end{figure}

\begin{figure}[ht]
\centering
  \includegraphics[width=90mm]{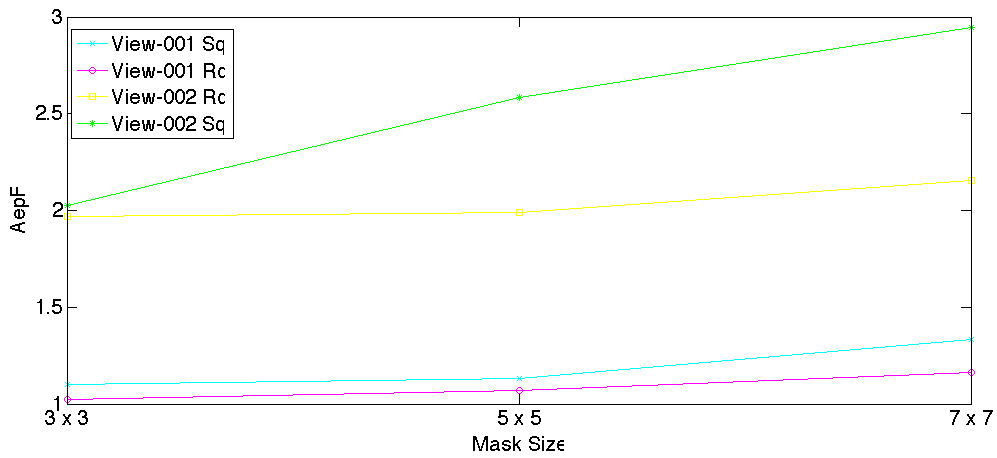}
\caption{Variation of the $AepF$ according to geometry and size of masks for View-001 and View-002 (S1\_L1\_Time13-57). }
\label{fig:MaskXAepF}
\end{figure}

\begin{figure}[htpb]
\centering
  \includegraphics[width=90mm]{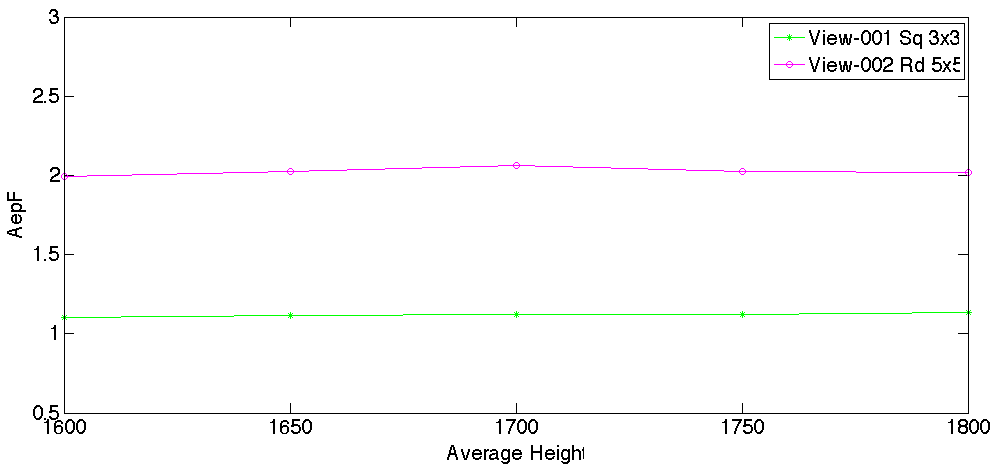}
\caption{Variation of the $AepF$ according to average person height for View-001 and View-002 (S1\_L1\_Time13-57). }
\label{fig:HeightXAepF}
\end{figure}

First, there are several configurations and parameters that could be adjusted for the corner point method and that have a direct impact on the performance. The main ones are: the number of regions at the ground plane; size and geometry of the mask that aggregates the neighbor corner points; average person height. Concerning the number of regions at the ground plane we have evaluated values between 1 and 40, where 1 means that we consider the same weight for all corner points detected in the view, while 40 corresponds that each region will have approximately one meter at the real world. Recalling that for each region we will have a weight that will be assigned to the corner points that fall into the region to compensate the distance of the person to the camera. Fig. \ref{fig:RXAepF} shows the impact of the number of regions on the $AepF$ considering a $3\times 3$ squared mask. The same behavior was observed for View-001 and different mask geometry and size. Therefore, we have fixed to 37 the number of regions, as shown in Fig. \ref{fig:CircularRegion}. The geometry of the masks to aggregate neighbor corner points are squared and radius while the size varies from $3\times 3$ to $7\times 7$. Fig. \ref{fig:MaskXAepF} shows the impact of the mask geometry and size on the $AepF$ considering 37 regions. The errors achieved using the circular masks are slightly lower than the square masks. Small masks are also better than the large ones. Finally, Fig. \ref{fig:HeightXAepF} shows that the different values for the average person height produce small changes to the $AepF$. For instance, if we do not consider the person height and do not correct the position of the projection corner points, the $AepF$ raises to values above 4. 

Tab. \ref{tab:1} shows the best results achieved on the video sequence S1\_L1\_Time13-59 while adjusting the parameters at video sequence S1\_L1\_Time13-57. The column Ground Truth stands for the ground truth being used: View\_001, View\_002 or Scene. The View\_001 ground truth only considers people who appear in View\_001, as well as the View\_002 ground truth only considers people who appear in View\_002. The scene ground truth considers people who appear at least in one of the two views. Tab. \ref{tab:1} also shows the results achieved by the method proposed by Albiol et al. \cite{Albiol2009}. The view column indicates if View\_001, View\_002 or both (Scene) were used for testing. The next two columns indicate if a mask or a radius was used for corner point extraction and its size (in pixels). The column combination tells us how the counting results from the two views were combined in our method (maximum, minimum and average value between the two views for each frame) as defined in Eq. \ref{eq:max}-\ref{eq:avg}. The last column shows the $AepF$ as defined in Eq. \ref{eq:perfmea}.
\begin{table*}
 \renewcommand{\arraystretch}{1.3}
 \small
 \caption{Results on video sequence S1\_L1\_Time13-59 while adjusting the parameters on video sequence S1\_L1\_Time13-57}
 \label{tab:1}
 \begin{center}
 \begin{tabular}{|c|c|c|c|c|c|c|c|}
 \hline
  Ground  & Average People & Method & Testing  & Mask/Radius & Size & Combination & AepF \\
   Truth & Height &  &  View & &  &  &  \\
 \hline
View\_001 & -- & Albiol et al. \cite{Albiol2009} & $V_1$ & Mask & 5x5 & -- & 1.87 \\
 & 1750 & Proposed & $V_1$ & Mask & 5x5 & -- & 1.51 \\
\hline
View\_002 & -- & Albiol et al. \cite{Albiol2009} & $V_2$ & Radius & 7x7 & -- & 3.24 \\
 & 1600 & Proposed & $V_2$ & Radius & 7x7 & -- & 2.61 \\
\hline
 & -- & Albiol et al. \cite{Albiol2009} & $V_1$ & Mask & 7x7 & -- & 3.39 \\
 Scene & -- & Albiol & $V_2$ & Radius & 3x3 & -- & 2.99 \\
 & 1750 & Proposed & Both & Mask & 5x5 & Average & 2.03 \\
 \hline
 \end{tabular}
 \end{center}
 \end{table*}

The improvement achieved in percentage from the standard method proposed by Albiol et al \cite{Albiol2009} is $19.3\%$, $19.4\%$ and $32.1\%$ for the View\_001, View\_002 and Scene respectively.

Tab. \ref{tab:2} shows the best results achieved on the video sequence S1\_L1\_Time13-59 while adjusting the parameters at video sequence S1\_L1\_Time13-57. The results of the proposed method are also compared to the results of the method proposed by Albiol et al \cite{Albiol2009}.  

\begin{table*}
 \renewcommand{\arraystretch}{1.3}
 \small
 \caption{Training with video 13\_59 and testing upon 13\_57}
 \label{tab:2}
 \begin{center}
 \begin{tabular}{|c|c|c|c|c|c|c|c|}
 \hline
    Ground Truth & Average People & Method & Testing  & Mask/Radius & Size & Combination & Average Error \\
   Truth & Height &  &  View & &  &  &  per Frame\\
 \hline
View 1 & -- & Albiol \cite{Albiol2009}& $V_1$ & Mask & 7x7 & -- & 1.51 \\
 & 1700 & Proposed & $V_1$ & Mask & 5x5 & -- & 1.05 (31.1\%) \\
\hline
View 2 & -- & Albiol \cite{Albiol2009}& $V_2$ & Radius & 5x5 & -- & 2.84 \\
 & 1600 & Proposed & $V_2$ & Radius & 5x5 & -- & 1.99 (29.7\%) \\
\hline
 & -- & Albiol \cite{Albiol2009}& $V_1$ & Mask & 7x7 & -- & 4.40 \\
 Scene & -- & Albiol & $V_2$ & Radius & 5x5 & -- & 5.09 \\
 & 1800 & Proposed & Both & Mask & 7x7 & Max & 2.34 (46.7\%) \\
 \hline
 \end{tabular}
 \end{center}
 \end{table*}

Tables \ref{tab:1} and \ref{tab:2} show that the proposed method outperforms the counting results of Albiol et al. \cite{Albiol2009} method by decreasing the average error per frame for both the views and the scene.

\subsection{Head Detection Results}
A dataset containing images of heads and non-heads was generated using two views of one video of the PETS2009 database. This dataset is used to train the head detection algorithms. For each view, we ended up with 2,100 images of heads and non-heads. The head images were normalized to size $9\times 9$ pixels. This size was chosen because of the variation on the size of the heads throughout the video, depending on the distance between the subjects and the camera. The next tables show the results for the SVM based head detection and the Haar Feature Adaboost Perceptron head detection. The parameters are average height and false alarm (for the Haar approach).

\subsubsection{Support Vector Machine}

\begin{table}
 \renewcommand{\arraystretch}{1.3}
  \small
 \caption{SMV test upon 13\_57}
 \label{tab:3}
 \begin{center}
 \begin{tabular}{|c|c|c|c|c|}
 \hline
 Average People  & \multicolumn{3}{c|}{ Average Error per Frame} \\
\cline{2-4}
  Height & View 1 & View 2 & Scene \\
 \hline
                  1600 & & &21.27\\
                  1650 & & &21.56\\
                  1700 & 3.05& 3.84 &21.99\\
                  1750 & & &22.53\\
                  1800 & & &23.03\\
 \hline
 \end{tabular}
 \end{center}
 \end{table}

\begin{table}
 \renewcommand{\arraystretch}{1.3}
  \small
 \caption{SMV test upon 13\_59}
 \label{tab:4}
 \begin{center}
 \begin{tabular}{|c|c|c|c|c|}
 \hline
  Average People  & \multicolumn{3}{c|}{ Average Error per Frame} \\
\cline{2-4}
  Height & View 1 & View 2 & Scene \\
 \hline
                  1600 & & &16.14\\
                  1650 & & &16.59\\
                  1700 & 3.12& 3.24 &16.86\\
                  1750 & & &17.16\\
                  1800 & & &17.59\\
 \hline
 \end{tabular}
 \end{center}
 \end{table}

The individual view count error is worse than the error achieved with the corner point method, but it is still acceptable. However, the difficulty in determining a good strategy to deal with correspondence of points of the two views impacts directly on the results for the scene count.

\subsubsection{Haar Feature based Adaboost Perceptron}
The cascading of the classifiers allows only the sub-images with the highest probability to be analyzed for all Haar-features that distinguish an object. It also allows one to vary the accuracy of a classifier. One can increase both the false alarm rate and positive hit rate by decreasing the number of stages. The inverse of this is also true.


For the Adaboost Perceptron classifier, besides the heads and non-heads dataset, 3,000 negatives images (in which there are no heads and their dimension is larger than 9x9, as shown in Figure \ref{fig:7} were used to train both views. These images are available in the Haar Training Tutorial \cite{Seo2012}. The Haar classifier was trained with a false alarm rate of 0.4 and 0.45.

\begin{table}
 \renewcommand{\arraystretch}{1.3}
  \small
 \caption{Adaboost Perceptron test upon 13\_57}
 \label{tab:5}
 \begin{center}
 \begin{tabular}{|c|c|c|c|c|c|}
 \hline
  Average People  & False Alarm & \multicolumn{3}{c|}{ Average Error per Frame} \\
\cline{3-5}
  Height &  & View 1 & View 2 & Scene \\
 \hline
                  1600 & & & & 21.83\\
                  1650 & & & & 22.08\\
                  1700 & 0.40& 8.19&7.96 &22.47\\
                  1750 & & & &22.96\\
                  1800 & & & &23.61\\
 \hline
                  1600 & & & & 19.40\\
                  1650 & & & & 19.82\\
                  1700 & 0.45& 3.86 & 14.05 & 20.42\\
                  1750 & & & & 21.24\\
                  1800 & & & & 22.00\\
\hline
\end{tabular}
 \end{center}
 \end{table}

\begin{table}
 \renewcommand{\arraystretch}{1.3}
  \small
 \caption{Adaboost Perceptron test upon 13\_59}
 \label{tab:6}
 \begin{center}
 \begin{tabular}{|c|c|c|c|c|c|}
 \hline
  Average People  & False Alarm & \multicolumn{3}{c|}{ Average Error per Frame} \\
\cline{3-5}
  Height &  & View 1 & View 2 & Scene \\
 \hline
                  1600 & & & & 18.86\\
                  1650 & & & & 18.86\\
                  1700 & 0.40& 11.59&11.53 &18.84\\
                  1750 & & & &18.87\\
                  1800 & & & &18.88\\
 \hline
                  1600 & & & & 17.73\\
                  1650 & & & & 17.88\\
                  1700 & 0.45& 6.98 & 4.54 & 17.92\\
                  1750 & & & & 17.88\\
                  1800 & & & & 18.08\\
\hline
\end{tabular}
 \end{center}
 \end{table}

With the Adaboost Perceptron, even the results of the views are not acceptable. In order to achieve a better result, a larger database and more time to train the algorithm with lower values of false alarm are needed.

\section{Conclusion}
\label{sec:con}
In this paper we have presented two novel approaches for people counting in crowded and open environments that combine the information gathered by multiple views. While one approach focused on detecting each individual and counting them, the other considered the crowd as a whole and applied statistical techniques to count the number of people. Also, different than most papers in this area, this work fuses the information from multiple views to count the number of people in the scene and not only on single images or views.

The proposed indirect method shows an improvement over the method it is based on. This is due to the inclusion of a weighting scheme to compensate for the distance of the persons to the camera. This method also showed good results when applied to scene people counting, improving the result over the view counting. For further improving, it is possible to change the combination technique to a more suitable and complex one, but this task is not trivial.

The proposed direct method has many more difficulties to deal with than the indirect one, and most of them do not have a straightforward solution. Therefore, the results of this method were not acceptable. In order to make this approach work on this kind of scenario, it is necessary to find a way to make a good correspondence of points, dealing with all mentioned problems. Another point that could bring a great contribution to this method is the inclusion of temporal information. For instance, detecting a new head in the middle of the crowd for some consecutive frames could mean that someone have been occluded until now and not that it is a false detection.

\begin{figure}[t]
    \centering
        \includegraphics[width=50mm]{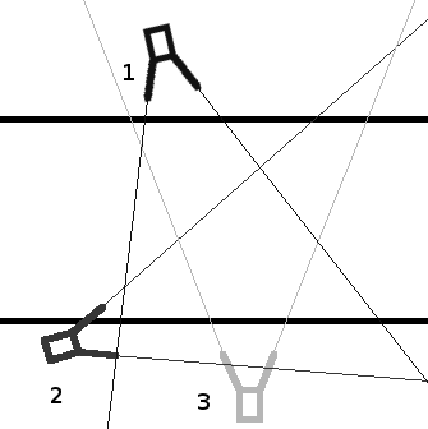}
    \caption{The cameras 1 and 3 are positioned ideally, while camera 2 can generate some problems for filming all the walkway.}
    \label{fig:9}
\end{figure}

There are other improvements that could be made to both methods, but that are not trivial, like changing the adjustment of the positions of the points of interest after applying the homography. During this work, it was possible to observe that the cameras' location is very important for the final counting result. If the camera is too high, the projections will be too small. And if the camera is too close to the ground plane, the projections will be too large. The height of the cameras used on PETS2009 is ideal (between 5 and 7 meters). Their position according to the walkway is also important. All cameras should capture only a part of the scene and not follow the path of the people, as exemplified on Figure \ref{fig:9}. View 2 of the PETS2009 database, unfortunately, is in this category. The main conclusion is that the indirect method performs much better than the direct method for this type of scenario. But there is still much work to do and improvements for both approaches are possible.

\bibliographystyle{IEEEtran}


\bibliography{counting-ESWA}

%



\end{document}